\documentclass[10pt,twocolumn,letterpaper]{article}

\usepackage{cvpr}  
\usepackage[accsupp]{axessibility}  % Improves PDF readability for those with disabilities.

\usepackage{algorithm}
\usepackage{algpseudocode}
\usepackage{amsmath}
\usepackage{multirow}

\newcommand{\myparagraph}[1]{\vspace{2pt}{\noindent\textbf{#1}}}

\DeclareMathOperator*{\TrimOff}{trim}

\definecolor{cvprblue}{rgb}{0.21,0.49,0.74}
\usepackage[pagebackref,breaklinks,colorlinks,allcolors=cvprblue]{hyperref}

\def\paperID{35785}
\def\confName{CVPR}
\def\confYear{2026}

\title{Bridging Domains through Subspace-Aware Model Merging}

\author{Levy Chaves$^{1}$, Chao Zhou$^{2}$, Rebekka Burkholz$^{2}$, Eduardo Valle$^{3}$, Sandra Avila$^{1}$\\
$^{1}$ Universidade Estadual de Campinas (UNICAMP), Recod.ai Lab., Instituto de Computação, Brasil\\
$^{2}$ CISPA Helmholtz Center for Information Security, Saarbrücken, Germany\\
$^{3}$ Intercom
}

\begin{document}
\maketitle
\begin{abstract}

Model merging integrates multiple task-specific models into a single consolidated one. Recent research has made progress in improving merging performance for in-distribution or multi-task scenarios, but domain generalization in model merging remains underexplored. We investigate how merging models fine-tuned on distinct domains affects generalization to unseen domains. Through an analysis of parameter competition in the task matrix using singular value decomposition, we show that merging models trained under different distribution shifts induces stronger conflicts between their subspaces compared to traditional multi-task settings. To mitigate this issue, we propose SCORE (Subspace COnflict-Resolving mErging), a method designed to alleviate such singular subspace conflicts. SCORE finds a shared orthogonal basis by computing the principal components of the concatenated leading singular vectors of all models. It then projects each task matrix into the shared basis, pruning off-diagonal components to remove conflicting singular directions. SCORE consistently outperforms, on average, existing model merging approaches in domain generalization settings across a variety of architectures and model scales, demonstrating its effectiveness and scalability.
\end{abstract}    
\section{Introduction}
\label{sec:intro}

% the motivation of model merging

The widespread availability of pre-trained models and public repositories has driven the development of techniques for efficiently combining and reusing existing models. Among these techniques, model merging~\cite{yang2024modelmergingllmsmllms} combines multiple fine-tuned models into a single unified model without requiring shared data or further fine-tuning. It directly merges the parameters of models fine-tuned from the same pre-trained backbone, enabling inference with a single forward pass, reducing both latency and storage. However, naive parameter averaging often performs poorly, as fine-tuning across tasks leads to divergent parameter distributions, resulting in interference and conflicts between task-specific representations. Addressing these inconsistencies is crucial for  effective and reliable model merging.

% how MM works and its characteristics, introducing task vectors
% Model merging (MM) is a technique designed to combine the strengths of multiple fine-tuned models into a single unified model without requiring data sharing or additional fine-tuning. It produces one model by directly merging the parameters of several fine-tuned models that share the same pretrained backbone. During inference, only a single forward pass through this merged model is needed, eliminating the need to call multiple models and thereby reducing both inference time and storage costs. However, naively averaging parameters across models typically performs poorly. Fine-tuning on different tasks often leads to divergent parameter distributions, and simple averaging can introduce interference or even conflicts between task-specific representations. Addressing these inconsistencies is crucial for developing effective and reliable model merging strategies.

% Different MM methods, from 
% 1. old-one like ties to,
% 2. svd decomposition

A popular strategy uses \textit{task vectors}~\cite{ilharcoediting2023}, defined as the difference between fine-tuned and pre-trained weights. TIES~\cite{yadav2024ties} identified redundancy in these vectors and proposed trimming small-magnitude elements while aggregating only those with consistent signs to mitigate conflicts. DARE~\cite{dare} introduced stochastic sparsification, showing that up to 99\% of model's parameters can be removed in large models without notable degradation. More recently, Task Singular Vectors (TSV)~\cite{tsv2025} applied singular value decomposition (SVD) to task matrices at the layer level to measure task interference and merge parameters within the singular subspace. 

% Despite significant progress in model merging, most existing works primarily evaluate in-distribution performance, while the impact of model merging on domain generalization remains largely unexplored. In this paper, we address this gap by exploring model merging as a compositional tool to leverage fine-tuned models across domains, where each domain represents different distribution shifts. By merging models from different distribution shifts, we expect the merged model’s generalization performance in an out-of-distribution domain to be better than in zero-shot evaluation. Concretely, we consider two models: one fine-tuned on images of animals in rocky terrains and another trained on animals in snowy environments. While each model performs well within its respective domain, their merged counterpart can potentially excel in an unseen domain such as animals in snowy mountain terrains, where both ``rocky" and ``winter" attributes \text{coexist}. 

Despite significant progress in model merging, most existing works primarily evaluate in-distribution performance, while the impact of model merging on domain generalization remains largely unexplored. Concretely, we treat each fine-tuned model as an expert on a particular distribution shift (a domain) and compose those experts to generalize to unseen domains. For intuition, consider two models fine-tuned on animals: one in rocky terrains, another in snowy land. Each model performs well in its domain, but neither captures both ``rocky'' and ``wintry'' attributes. By merging these models, we expect the merged model to combine complementary domain-specific representations and therefore to perform better on an unseen target such as animals in snowy mountain terrains, where these attributes coexist.

Analyzing the singular subspaces of such models, we observe subspace conflicts between fine-tunings on different domains that are much stronger than those of standard multi-task settings. We argue that such conflicts is challenging for SVD-based model merging and, to mitigate this issue, we propose SCORE, a Subspace COnflict-Resolving mErging method.

Our main contributions are:
\begin{enumerate}
\item We study the domain generalization  of merged models using a leave-one-domain-out protocol.
\item We show that the overlap of task-matrix singular subspaces under domain generalization is significantly higher than under multi-task \text{scenarios}.
\item We propose SCORE, a Subspace COnflict-Resolving mErging method that  improves domain generalization.
\item Our extensive experiments show that SCORE consistently outperforms, on average, existing model merging methods across eight domain generalization benchmarks and three model scales.
\end{enumerate}
\section{Related Work}
\label{sec:related-work}

% Literature overview 

\paragraph{Model merging}\hspace{-0.2cm}offers an efficient opportunity for combining multiple models without retraining.  Several approaches rely on neuron permutations to align the models into a shared optimization basin through weight permutation strategies~\cite{ainsworthgit2023,stoicazipit2024,xu2024training}, or activations~\cite{jordan2023repair}. Another avenue focuses on the multi-task scenario, where a single pre-trained model is individually fine-tuned for different tasks and then combined for a multi-task model.   \text{\citet{ilharcoediting2023}} introduced the concept of \textit{task vectors}, which are the parameter-wise difference between fine-tuned  and pre-trained models, and the goal is to merge all available task vectors to create a multi-task model. TIES~\cite{yadav2024ties} addressed parameter redundancies, \ie, when parameters benefit one task but not another, by first selecting the \text{top-k} most significant parameter changes and then constructing a sign vector based on the majority sign across all models. The latter is used to merge the task vectors disjointly, meaning the average is not computed when a parameter is zero or when parameters disagree in sign. Other alternatives to account for parameter redundancy include parameter sign consensus~\cite{wang2024localizing}, randomly dropping and rescaling remaining parameters~\cite{dare}, keeping only the middle magnitudes~\cite{davari2024model}, keeping only the highest magnitudes~\cite{marczak2024magmax}, or inter and cross-parameter competition to adjust the scale of individual parameters~\cite{pcb}. 

% Svd for merging (some examples )
Another line of work mitigates parameter interference using Singular Value Decomposition (SVD) of individual weight matrices. KnOTs~\cite{stoicamodel2025} concatenated task-specific low-rank adapters and averages right-singular vectors before SVD reconstruction, while Twin Merging~\cite{lu2024twin} employs a router network to select among SVD experts. TSV~\cite{tsv2025} combined multiple SVD decompositions and orthogonalized the resulting singular vectors to reduce interference. ISO-C~\cite{isoc2025} enforced a uniform (isotropic) spectrum, improving multi-task performance. 

% Training 
Other methods rely on gradient-based optimization or validation-set information. Fisher Merging~\cite{fishermerging} and RegMean~\cite{regmean} performed weighted averaging using the Fisher information matrix and input vector inner products, respectively. \citet{ortiz2023task} demonstrated the benefits of NTK linearization for merging, while \citet{AdaMerging_ICLR_2024} learns layer-wise scalers through test-time training. EMR-Merging~\cite{huang2024emr} further introduces per-task parameter masks and rescalers, assuming access to target fine-tuned weights.

\paragraph{Model merging for generalization.} 

\citet{wortsman2022model} showed that averaging parameters of multiple ImageNet fine-tuned models with different hyperparameter configurations can improve robustness against distribution shifts. \citet{rame2023model} improved previous work by requiring access to auxiliary tasks related to the target one, and relying on a two-step fine-tuning to recycle public models available for similar data as auxiliary tasks.  While these methods show improvements, they rely on averaging strategies across hyper-parameters configurations or require access to validation datasets for hyperparameter selection. 

For strategies beyond model averaging, \citet{AdaMerging_ICLR_2024} also considered corrupted versions of ImageNet as out-of-distribution evaluation, but their method relies on accessing test data samples to perform costly test-time training. However, as our scope is to evaluate the merged model without accessing any data or any optimization, our method only requires access to the fine-tuned source models and does not require any further training or gradient steps. 

\citet{pcb} adopted a fixed set of generalization tasks for emotion classification in NLP tasks, without any further variation of such out-of-distribution datasets. Practitioners expect model-merging methods to be robust to several distribution shifts, and using a fixed set of datasets is suboptimal because we are only measuring performance for that particular case, making it unclear whether these methods generalize to other out-of-distribution data.  

Current evaluation frameworks either compromise the advantages of model merging through data access requirements~\cite{rame2023model,AdaMerging_ICLR_2024,tam2024realistic} or provide insufficient evidence of generalization across a fixed set of generalization tasks~\cite{pcb}. In contrast, our work addresses these concerns by requiring only access to fine-tuned source models and by considering merging models without any additional training, optimization, or privileged information. To the best of our knowledge, we are the first to investigate model merging under the domain generalization scheme of leave-one-out evaluation, providing a comprehensive evaluation across multiple distribution-shift scenarios.

\section{Background and Motivation}
\label{sec:backgorund}

In this section, we first outline the general framework for model merging and domain generalization, explaining how we connect these two concepts. Next, we introduce the notation used throughout the paper. Finally, we motivate our approach by measuring the subspace overlap between fine-tuned models across different domains, demonstrating that this overlap is greater than what is typically observed in traditional multi-task settings.

\subsection{Preliminaries}
\label{subsec:preliminaries}

\paragraph{Model merging.} Given a collection of $D$ fine-tuned model weights $\{\theta_{1}, \theta_{2}, ..., \theta_{D}\}$ from the same pre-trained model architecture $\theta_{pre}$, we first compute the set of delta weights $\Delta w = \{\Delta{w}_d \mid \Delta{w}_d:= \theta_{d} - \theta_{\text{pre}}\}_{d=1}^{D}$, \ie, the parameter-wise difference between the fine-tuned and pre-trained model and then merge all delta weights using a merging function $f_{\text{merge}}$, such that $\Delta_{\text{merged}} = f_{\text{merge}}(\Delta w, \lambda)$. Here, $f_{\text{merge}}$ represents a merging method from the literature (\eg, Section~\ref{sec:related-work}), and $\lambda \in \mathbb{R}$ is a scaling factor. After that, we sum $\Delta_{\text{merged}}$ to the pre-trained model as $\theta_{\text{final}} = \theta_{\text{pre}} + \Delta_{\text{merged}}$.

\paragraph{Domain generalization.} In this work, we consider a scheme similar to multi-source Domain Generalization (DG) setting~\cite{zhou2022domain,nicopp}, where we are given $D$ distinct source domains $\mathcal{S} = \{S^{d}\}_{d=1}^D$. A domain $d$ is defined by a joint distribution $P_{XY}^{d}$ over a domain-specific input space $\mathcal{X}^{d}$ and a shared target space $\mathcal{Y}$. Each source domain $S^{d}$ is associated with a unique distribution $P_{XY}^{d}$, such that $P_{XY}^{d} \neq P_{XY}^{d'}$ for $d \neq d'$. The objective is to learn a single predictive model $f: \mathcal{X} \to \mathcal{Y}$ using only the data from $\mathcal{S}$. This model must generalize to an unseen target domain $T$ associated with a novel distribution $P_{XY}^T$, where $T \notin \mathcal{S}$. Typical approaches run a single training containing all source-domain data and rely on a domain-generalization strategy~\cite{robey2021model,zhang2022towards} or specifically designed loss function~\cite{liang2025comprehensive,krueger2021out,gulrajanisearch2021}.

\paragraph{Model merging for domain generalization.} We introduce a novel perspective on domain generalization through model merging as a compositional mechanism that enables better generalization to unseen domains. Our study investigates whether merging models fine-tuned on data from different distribution shifts can result in a single model with improved generalization to previously unseen data. This setting mirrors practical use cases of generalization, such as combining distinct visual styles or object concepts in generative models, or enabling zero-shot transfer to new tasks.

\begin{figure}
    \centering
    \includegraphics[width=\linewidth]{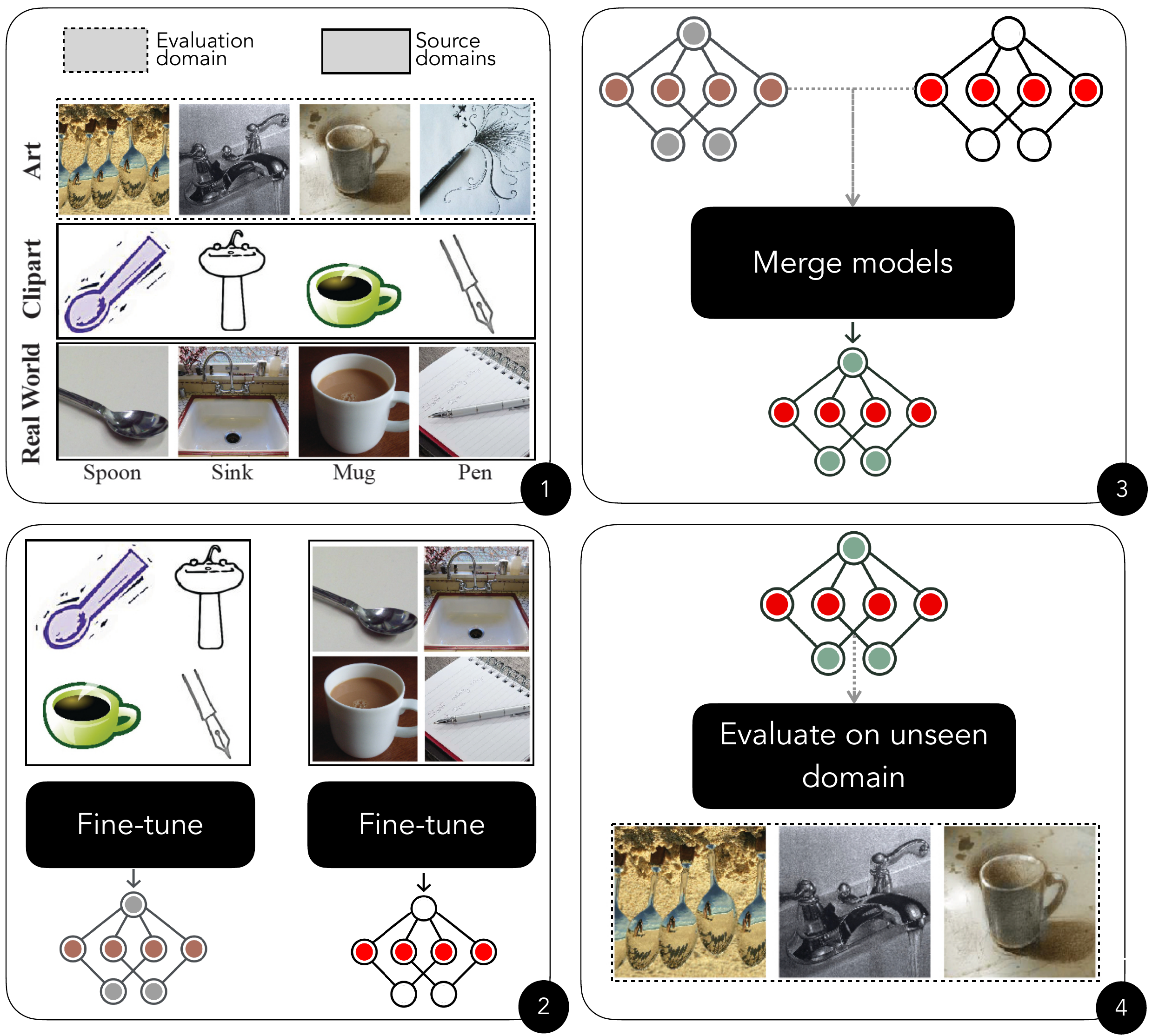}
    \caption{Evaluating model merging for domain generalization. We merge fine-tuned models across multiple domains and evaluate the merged model on an unseen domain. We repeat the leave-one-out process for each available domain.}
    \label{fig:MM OOD protocl}
\end{figure}

We evaluate merging in a leave-one-domain-out protocol~\cite{gulrajanisearch2021,zhou2022domain,chavesweight2025}: given \(D\) domains \(\{S^{d}\}_{d=1}^{D}\) we hold out a domain \(S^{t}\) as the unseen target and use the remaining \(D-1\) domains as sources. For each source domain \(S^{d}\) we assume access only to a fine-tuned parameter checkpoint \(\theta_d\) (obtained by fine-tuning a common pre-trained backbone \(\theta_{\text{pre}}\) on \(S^{d}\)), while source training data is unavailable. We form the set of delta parameters \(\Delta w = \{\Delta{w}_d\}_{d\neq t}\) with \(\Delta{w}_d=\theta_d-\theta_{\text{pre}}\) and compute a merged delta as: 
\[
\Delta_{\text{merged}} \;=\; f_{\text{merge}}(\{\Delta{w}_d\}_{d\neq t};\; \lambda),
\]

\noindent where \(f_{\text{merge}}\) is a data-free and optimization-free model merging method. The hyper-parameter $\lambda$ controls the contribution of the merged weights to the pre-trained model. Prior work often tunes $\lambda$ using validation data~\cite{ilharcoediting2023,pcb,wang2024localizing,marczak2024magmax}, but in our setting, accessing the out-of-distribution data is prohibited, so we fix $\lambda=1$. We evaluate the resulting model \(\theta_{\text{final}}=\theta_{\text{pre}}+\Delta_{\text{merged}}\) on the test set of \(S^t\). We repeat this procedure for each available domain \(d=1,\dots,D\). Figure~\ref{fig:MM OOD protocl} shows the protocol for a single evaluation domain.

\subsection{Motivation}
\label{sec:motivation}

% We explore model merging as a compositional tool to leverage fine-tuned models across domains with distribution shifts. By integrating knowledge from these different distribution shifts, we aim to enhance the merged model’s generalization performance in an out-of-distribution domain. Concretely, consider two models: one fine-tuned on images of animals in rocky terrains and another trained on animals in snowy environments. While each model performs well within its respective domain, their merged counterpart can potentially excel in an unseen domain such as animals in snowy mountain terrains, where both ``rocky" and ``winter" attributes coexist. 

In traditional multi-task learning, the evaluation task is part of the set of models being merged, and the goal is to preserve information from all tasks so the merged model performs well across them. These tasks are diverse and different both in label space and in semantics, \eg MNIST (digit classification) and RESIC45 (landscape classification). 

On the other hand, domain generalization presents a more challenging scenario for model merging. Models trained on different domains often share the same label space but differ in their data distributions, encouraging each model to learn class-discriminative and domain-specific representations. Because these models solve similar classification problems, their delta weights tend to align along similar singular directions. This strong subspace overlap contrasts sharply with the diversity of multi-task learning. This overlap leads to interference among singular directions during merging, as competing dominant directions may translate into competing features in the merged model. Effectively handling this conflict becomes crucial for improving generalization in unseen domains.

% Consequently, merging must resolve conflicts between correlated but distribution-shifted representations. 

To quantify the overlap between subspaces, we use the \textit{Subspace Alignment Ratio} (SAR) metric~\cite{isoc2025}. The SAR metric between two delta weights $\Delta w_{i}$ and $\Delta w_{j}$  is defined~as:

\begin{equation}
\mathrm{SAR}(\Delta w_i, \Delta w_j; k_j) = 
\frac{\lVert \Pi_{k_j, j} \, \Delta w_i \rVert_F}
     {\lVert \Delta w_i \rVert_F},
\end{equation}

\noindent where $\Pi_{k_j, j} = U_{k_j, j} U_{k_j, j}^\top$ is the projection matrix 
onto the subspace spanned by the top $k_j$ left-singular vectors of 
$\Delta w_j$. The columns of $U_{k_j, j}$ are obtained from the SVD 
decomposition of $\Delta w_j$, and the number of singular vectors $k_j$ is determined from $\Delta w_j$ by minimizing the approximation error $\epsilon$:
\begin{align}
\label{eq:sar}
k_j 
&= \min \left\{ k : 
\lVert \Delta w_j - \Pi_{k_j, j} \, \Delta w_j \rVert_F 
\leq \epsilon \, \lVert \Delta w_j \rVert_F \right\} \notag \\
&= \min \left\{ k : 
\frac{\sum_{r=i=k+1}^{r} \sigma_i^2}{\sum_{i=1}^{r} \sigma_i^2} 
\leq \epsilon^2 \right\},
\end{align}

\noindent where $\Sigma = \mathrm{diag}(\sigma_1, \ldots, \sigma_r)$ contains the 
singular values of $\Delta w_j$, and the equivalence follows from the 
definition of the Frobenius norm.

The SAR quantifies the alignment between the subspaces of two delta matrices as a function of the number of dominant singular vectors of $\Delta w_j$. We denote $\mathrm{SAR}_{\mathrm{avg}}$ the 
\textit{Average Subspace Alignment Ratio} across all layers to provide a single score measuring the overlap between two models.

\begin{figure}[htb!]
  \centering
  
  % First subfigure
  \begin{subfigure}[t]{0.49\columnwidth}
    \centering
    \includegraphics[width=\linewidth]{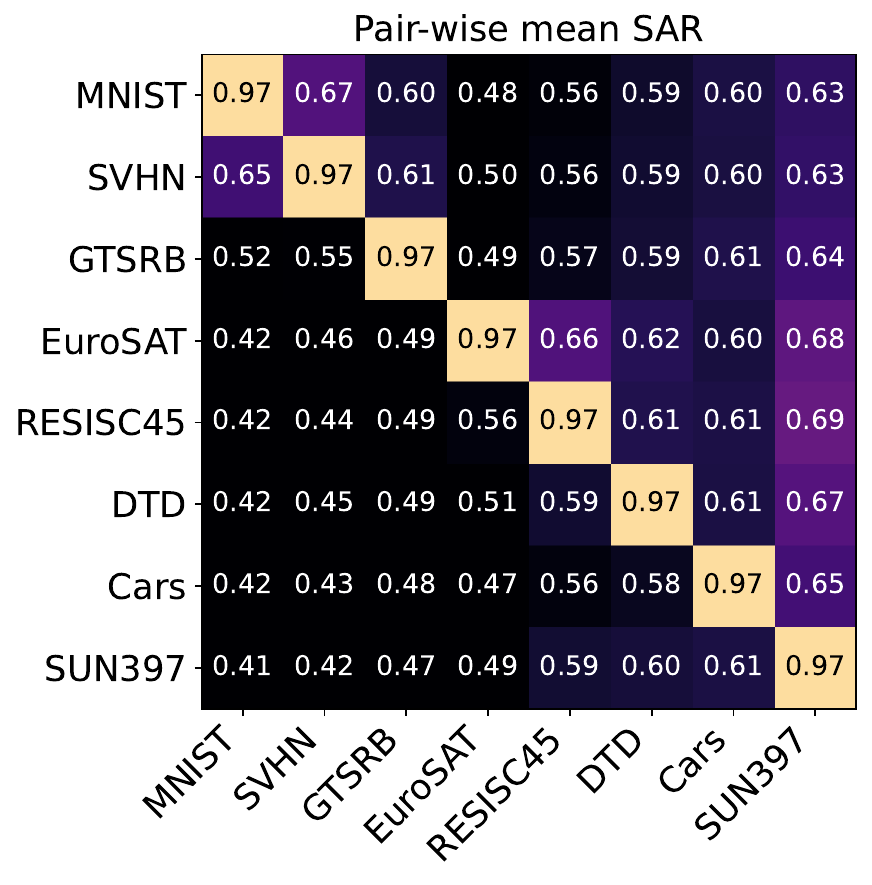}
    \caption{Multi-task learning}
    \label{fig:case1}
  \end{subfigure}
  \hfill
  % Second subfigure
  \begin{subfigure}[t]{0.49\columnwidth}
    \centering
    \includegraphics[width=\linewidth]{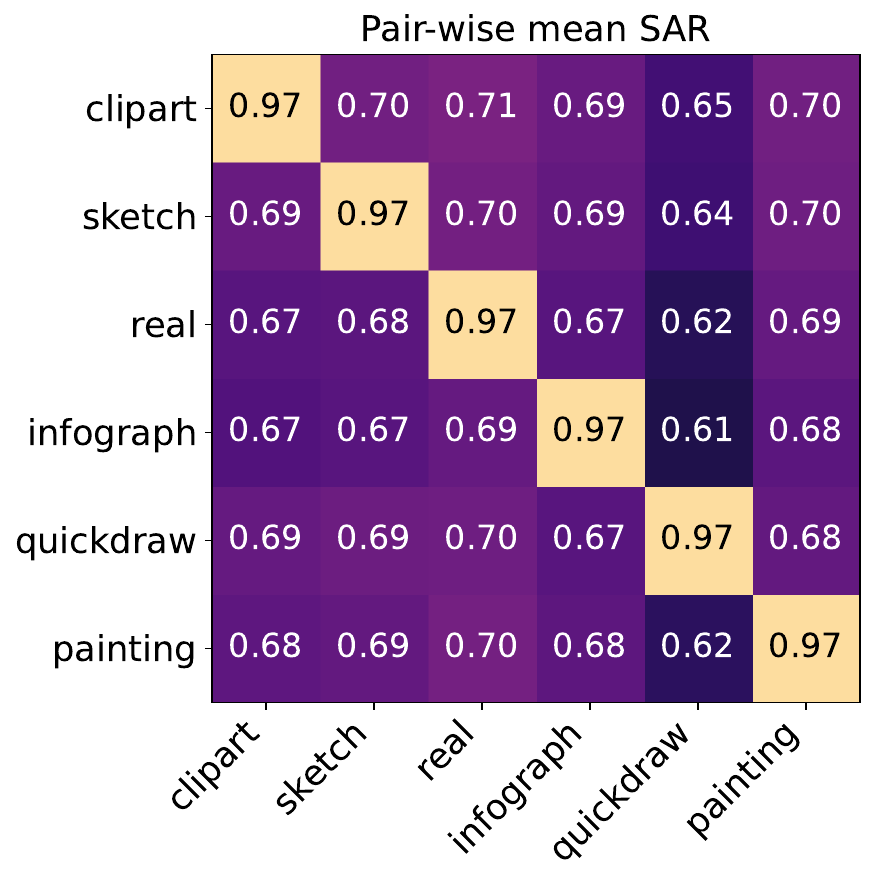}
    \caption{Domain generalization}
    \label{fig:case2}
  \end{subfigure}
  
  % Main caption for both figures
  \caption{Pairwise similarity of fine-tuned models measured by the Subspace Alignment Ratio (SAR). (a) Models fine-tuned for 8~datasets from \citet{ilharcoediting2023}. (b) Models fine-tuned for 6 domains of DomainNet~\cite{domainnet}. The similarity between multi-domain models is much higher than between multi-task models, with much more overlap between the subspaces occupied, creating opportunity for conflicts.}
  \label{fig:mean-sar}
\end{figure}

Figure~\ref{fig:mean-sar} shows the average SAR metric between tasks for merging 8 datasets in a multi-task setting~\cite{ilharcoediting2023}, and between domains in domain generalization for the 6 domains in the DomainNet dataset. In multi-task learning, the overlap is more pronounced among groups of related tasks, such as MNIST, SVHN (digit classification), and EuroSAT, RESIC45 (landscape classification). In contrast, domain generalization produces consistently higher SAR values, even though all domains correspond to the same underlying task but differ in data distribution. We also report an additional overlap measure based on principal angles~\cite{subspace-method} in Appendix Section~\ref{supp:subsec-overlap}

\vspace{-0.3cm}

We argue that domain shifts induce subspaces that overlap more strongly than task shifts. Such overlap introduces additional challenges for SVD-based merging methods. These methods prioritize the top singular directions and their corresponding eigenvalues. However, when fine-tuned models across different domains share similar singular directions, conflicts can arise during inference. In practice, these competing directions from one domain can dominate those from another, potentially biasing the merging process toward tasks with larger singular values and, thus, harming out-of-distribution generalization performance.

% On the other hand, domain generalization presents higher SAE than the multi-task setting, in which each domain represents the same target tasks, but different in data distribution. We argue that this phenomenon presents an extra challenge for SVD-based merging methods, since these methods often focus on the top singular directions and their corresponding eigenvalues. For instance, if the singular directions specific to different tasks are very similar, but each task disagree in their magnitudes in singular values space, leading to a conflicts during inference. This conflict arises because two (or more) singular directions might compete at different magnitudes.

\section{SCORE}
\label{sec:method}

In Section~\ref{sec:motivation}, we showed that there is substantial overlap between subspaces in domain generalization, which translates to higher conflicting singular directions and feature competition in the merged model. To address this issue, we propose SCORE, a Subspace COnflict-Resolving mErging method designed to alleviate such singular subspace conflicts. SCORE constructs a shared orthogonal basis by computing the principal components of the concatenated leading singular vectors of all models fine-tuned under domain shift. Each task matrix is then projected into this shared basis, where off-diagonal components shows the inter-domain conflicts in singular directions. We outline our method in Algorithm~\ref{algo:proposal} for a single layer, and describe each step in the following. 

\begin{algorithm}
\caption{SCORE} 
\label{algo:proposal}
\begin{algorithmic}[1] 
\Require Pre-trained model $\theta_{pre}$, Task matrices $\{\Delta_1, \dots, \Delta_D\}$,  number of domains to merge $D$

\Ensure Merged task matrix $\hat{M}$

\For{$d = 1$ to $D$}

\State Compute SVD: $\Delta_d = U_d \Sigma_d V_d^\top$ 

\State Retain first $\tfrac{1}{D}$ singular components of $U_d,V_d$

\EndFor 

\State \textbf{Concatenate} the matrices:
\[
U_{*} \leftarrow [\,U_1 \mid U_2 \mid \dots \mid U_D\,],~
V_{*} \leftarrow [\,V_1 \mid V_2 \mid \dots \mid V_D\,]
\]

\State Compute SVD of $U_{*}$: $ P_{U_{*}} \Sigma_{U_{*}} Q_{U_{*}}^\top$ 

\State Compute SVD of $V_{*}$: $ P_{V_{*}} \Sigma_{V_{*}} Q_{V_{*}}^\top$ 

\State $U_\perp \leftarrow P_{U_{*}} Q_{U_{*}}^\top$ \Comment{Find orthogonal basis for ${U_{*}}$} 

\State $V_\perp \leftarrow P_{V_{*}} Q_{V_{*}}^\top$ 
\Comment{Find orthogonal basis for ${V_{*}}$} 

\State $\Sigma_{score} \leftarrow \mathbf{0}$ 
\For{$d = 1$ to $D$} 
\State $\Delta'_d \leftarrow U_\perp^\top \Delta_d V_\perp$ 
\Comment{Change basis}

\State $\Sigma_{score} \leftarrow \Sigma_{score} + \TrimOff(\Delta'_d)$
\Comment{Eq.~\ref{eq-trim}} 

\EndFor

\State $\hat{M} \leftarrow U_\perp \Sigma_{score} V_\perp^\top$ \Comment{Reconstruct merged matrix} 
\State \Return $\theta_{pre} + \hat{M}$

\end{algorithmic} 
\end{algorithm}

\paragraph{Combining domain-specific directions.} Consider a collection of delta weights for the $l$-th network layer $\Delta w^{(l)} = \{\Delta w_{d}^{(l)}\}_{d=1}^{D}$ where each  $\Delta w_{d}^{(l)}=\theta_{d}^{(l)} - \theta_{pre}^{(l)} \in \mathbb{R}^{m \times n}$ represents the parameter difference for domain $d$ in layer $l$. Applying Singular Value Decomposition (SVD) to $\Delta w_{d}^{(l)}=U\Sigma~V^{T}$, where $U \in \mathbb{R}^{m \times r}$, $V \in \mathbb{R}^{n \times r}$ are the left and right singular vectors, respectively, and $\Sigma \in \mathbb{R}^{r \times r}$ is a diagonal matrix containing the singular values. For clarity, we omit the layer superscript $l$ in the following discussion and describe the merging process for a single layer.

To accommodate the singular directions of each individual domain, we compute the SVD decomposition of each delta matrix and consider only the top left and right singular directions: 
\begin{align*}
U_{*} \leftarrow [\,U_1 \mid U_2 \mid \dots \mid U_D\,], V_{*} \leftarrow [\,V_1 \mid V_2 \mid \dots \mid V_D\,],
\end{align*}

\noindent where $U_{*} \in \mathbb{R}^{m \times r}$ and $V_{*} \in \mathbb{R}^{n \times r}$ are the concatenated top-k domain-specific left and right singular directions from the SVD decomposition of $\Delta w_{d}$, respectively.  

\paragraph{Orthogonalization.} There is no guarantee that \( U_{*} \) and \( V_{*} \) are orthogonal to each other, which can compromise the reconstruction property of the SVD if used as they are. To create a shared subspace across domains that sticks to SVD constraints, we need to orthogonalize \( U_{*} \) and \( V_{*} \). Following~\citet{tsv2025}, we compute the SVD of $U_{*} = P_{U_{*}} \Sigma_{U_{*}} Q_{U_{*}}^{T}$ and $V_{*}= P_{V_{*}} \Sigma_{V_{*}} Q_{V_{*}}^{T}$ :
\begin{equation}
    U_\perp \leftarrow P_{U_{*}} Q_{U_{*}}^\top, V_\perp \leftarrow P_{V_{*}} Q_{V_{*}}^\top.
\end{equation}

This step yields $U_\perp$ and $V_\perp$, which form a shared input and output basis that is closest to all $D$ domain-specific subspaces.

\paragraph{Isolating subspace conflicts.} Our intuition is that if the domain-specific top singular directions are mostly orthogonal to each other, both  ($U_{\perp}$, and $V_{\perp}$) defines a representative shared basis, with minimal level of conflicting dominant directions. However, as the subspace overlap is not negligible in domain generalization (see Section~\ref{sec:motivation}), we need to quantify which dominant directions are conflicting and their respective magnitude. For this, we apply the change of basis for each domain-specific matrix using $U_{\perp}$ and $V_{\perp}$ to understand which dominant directions each domain-specific weights $\Delta_d$ use: 
\begin{align*}
\Delta'_d = U_\perp^\top \Delta_d V_\perp,
\end{align*}

\noindent where $\Delta'_{d}~\in\mathbb{R}^{r~\times~r}$.  The conflict in singular directions becomes explicit by reformulating the domain-specific SVD $\Delta_{d} = U_{d}\Sigma_{d}V_{d}$ with the shared projection:
$$
\Delta'_d = U_\perp^\top (U_d \Sigma_d V_{d}^\top) V_\perp= (U_\perp^\top U_d) \Sigma_d (V_{d}^\top V_\perp).
$$

Let us define $R_U^{(d)} = U_\perp^\top U_d$ and $R_V^{(d)} = V_d^\top V_\perp$. These matrices represent the dot products between the shared basis vectors and the domain $d$'s specific basis vectors. The change of basis isolates the conflicts in dominant singular directions by transforming the matrix so that we can clearly distinguish between two types of information:
\begin{enumerate}
    \item \textbf{Agreement (Diagonal):} The diagonal elements $(\Delta'_d)_{ii}$ measure the magnitude of domain $d$ along the $i$-th \textit{shared} principal direction. 
    \item \textbf{Conflict (Off-Diagonal):} The off-diagonal elements $(\Delta'_d)_{ij}$, for $i \neq j$ captures how domain $d$ couples the $i$-th and $j$-th shared directions. Large off-diagonals indicate cross-talk between shared coordinates.
\end{enumerate}

\begin{figure*}[htb!]
    \centering
    \includegraphics[width=0.99\linewidth]{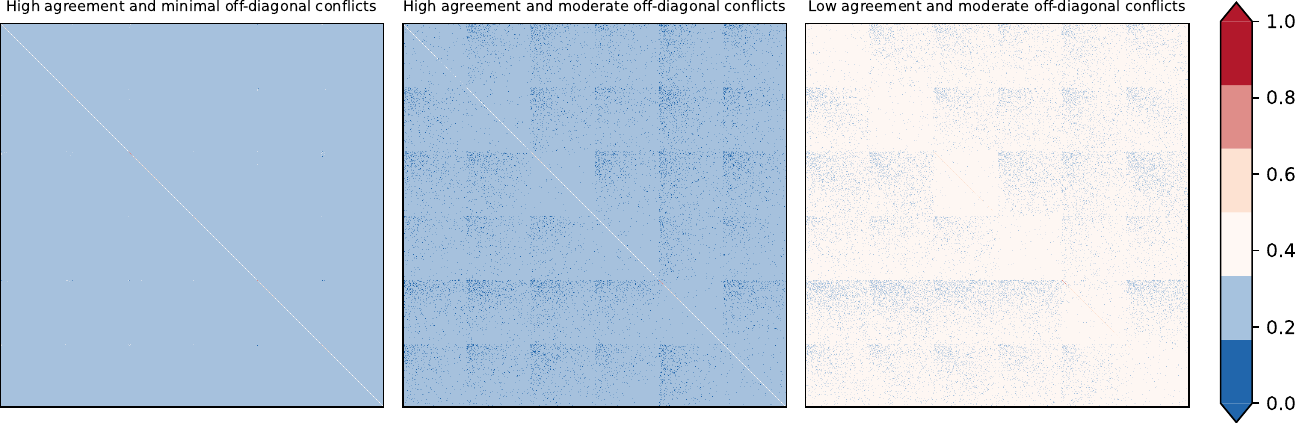}
    \caption{$\Sigma_{score}$ (see Alg.~\ref{algo:proposal}) computed over 6 DomainNet domains across different attention layers of a ViT-B-32. Left: high agreement/low conflict, with a strong main diagonal and clean off-diagonal. Middle: high agreement/high conflict, with both the main diagonal and off-diagonal blocks significantly occupied. Right: low agreement/hight conflict, with off-diagonal elements dominating and main diagonal nearly absent, indicating that the shared basis ($U_{\perp}$ and $V_{\perp}$) fails to capture consistent singular directions across domains and that those domains intensely compete for subspaces in the shared representation.}
    \label{fig:block-structure}
\end{figure*}

Figure~\ref{fig:block-structure} illustrates how block structures emerge after changing the basis for three distinct attention blocks in the ViT-B-32 model. For instance, we consider 6 DomainNet domains and then calculate \(\Sigma_{score} = \sum_{d=1}^{D=6} U_{\perp}^{T} \Delta_{d} V_{\perp}\). We present three scenarios, from left to right: 
\begin{itemize}
    \item \textbf{High agreement and minimal off-diagonal conflicts}: most energy lies on the diagonal, indicating the shared basis captures the dominant domain directions and domains align with the principal coordinates;
    \item \textbf{High agreement and moderate off-diagonal conflicts}: diagonal entries still dominate the magnitudes, but noticeable off-diagonal coupling appears, suggesting conflicts between shared directions;
    \item \textbf{Low agreement and moderate off-diagonal conflicts}: main diagonal energy is weak, and cross-terms are evident, indicating the shared basis failed to capture many domain-specific top directions. 
\end{itemize}

\paragraph{Trimming off-diagonal outliers.} Building on the previous analysis, we intend to keep the main diagonal elements of $\Delta'_{d}$, which capture the agreement between each domain and the shared principal directions. However, keeping only the diagonal entries may miss important off-diagonal components that carry significant information about shared variability across domains. Therefore, instead of discarding all off-diagonal terms, we retain those with significant magnitude while suppressing outlier or noisy couplings. 
%This strategy prevents the loss of informative cross-direction interactions that could otherwise degrade the final model performance.

Finally, to consolidate multiple  $\Delta'_{d}$ and reconstruct the merged matrix as $\hat{M} = U_{\perp}\Sigma_{score}V_{\perp}^{T}$, we compute the final directions   $\Sigma_{\mathrm{score}} = \sum_{d=1}^D \mathrm{trim}(\Delta'_d)$, where the $trim$ can be described as:
\begin{equation}
\resizebox{\columnwidth}{!}{%
  $\displaystyle
  \text{trim}(\Delta_k')_{ij} =
  \begin{cases}
  (\Delta_k')_{ii}, & \text{if } i = j \quad (\text{Keep the diagonal}) \\
  (\Delta_k')_{ij}, & \text{if } i \neq j \text{ and } |(\Delta_k')_{ij} - \mu_{\text{off}}| < \tau \cdot \sigma_{\text{off}} \\
  0, & \text{otherwise} \quad (\text{Prune outliers}),
  \end{cases}
  $
}
\label{eq-trim}
\end{equation}

\noindent where $\mu_{\text{off}}$ and $\sigma_{\text{off}}$ are the mean and standard deviation of all off-diagonal elements. Considering only the main diagonal indicates taking account of dominant directions; on the other hand, keeping the whole $\Delta'_{d}$ account for all pair-wise interferences may introduce high levels of noise in the final and merged weights.  We set $\tau = 1.96$ as it corresponds to 95\% confidence interval in a standard normal distribution.

\section{Experiments}
\label{sec:experiments}

\begin{table*}[htb!]
\centering
\scriptsize
\begin{tabular}{llccccccccc} % 11 columns total
\toprule
% First header cell is empty
 & \textbf{Method} & \textbf{PACS} & \textbf{DomainNet} & \textbf{ImageNetR} & \textbf{NICO++} & \textbf{OfficeHome} & \textbf{TerraIncognita} & \textbf{FedISIC} & \textbf{RetinaDomains} & \textbf{Avg} \\
\midrule
\midrule
%--- ViT-B-32 Section ---
% 10 rows, so \multirow{10}
\multirow{10}{*}{\rotatebox{90}{\textbf{ViT-B-32}}}
% This section now has 11 columns of data (1 'l' and 10 'c')
& Zeroshot & 94.85 & 56.10 & 66.91 & 77.14 & 79.82 & 22.73 & 23.08 & 24.40 & 55.63 \\
& Task. Arithm. & 93.40 & 5.40 & 4.65 & 56.92 & 66.73 & 40.47 & 32.18 & \underline{40.00} & 42.47 \\
& TIES & \underline{95.77} & 53.22 & 72.98 & 78.18 & 83.27 & \underline{47.03} & \textbf{50.42} & 37.84 & 64.84 \\
& MagMax & 95.46 & 43.87 & 68.38 & 76.18 & 80.37 & 43.56 & 43.32 & \textbf{41.06} & 61.53 \\
& PCB & 95.35 & \underline{58.53} & \underline{74.67} & \underline{80.19} & 83.59 & 40.84 & 38.41 & 35.45 & 63.38 \\
& TSV & \textbf{95.83} & 55.64 & 74.23 & 79.78 & \underline{83.74} & \textbf{47.68} & 44.22 & 38.44 & \underline{64.95} \\
& Iso-C & 95.37 & 55.84 & \textbf{74.74} & 79.81 & 82.34 & 36.53 & 32.91 & 32.97 & 61.31 \\
& Iso-CTS & 95.30 & 56.80 & 73.46 & 79.76 & 82.00 & 32.87 & 29.63 & 33.61 & 60.43 \\
& \textbf{SCORE (Ours)} & 95.70 & \textbf{58.68} & 73.44 & \textbf{80.59} & \textbf{84.23} & 46.92 & \underline{50.12} & 35.87 & \textbf{65.69} \\
& Task Experts & 99.81 & 74.43 & 86.70 & 84.34 & 91.65 & 82.15 & 61.26 & 48.72 & 78.63 \\
\midrule[1pt] % Thicker rule to separate the sections
%--- ViT-L-14 Section ---
% 10 rows, so \multirow{10}
\multirow{10}{*}{\rotatebox{90}{\textbf{ViT-L-14}}}
& Zeroshot & 98.39 & \underline{66.25} & 86.51 & 84.06 & 88.82 & 46.46 & 19.77 & 26.00 & 64.53 \\
& Task. Arithm. & 98.53 & 30.61 & 13.07 & 73.39 & 89.97 & 51.11 & 31.64 & 42.81 & 53.89 \\
& TIES & \underline{98.60} & 65.80 & 90.16 & \underline{85.53} & 91.26 & \underline{59.72} & 47.48 & 41.11 & \underline{72.46} \\
& MagMax & \textbf{98.62} & 62.24 & 88.15 & 84.82 & \textbf{91.41} & 55.16 & \textbf{51.69} & \underline{40.74} & 71.60 \\
& PCB & 98.54 & \textbf{68.26} & 90.38 & 86.44 & 91.13 & 59.32 & 38.93 & 38.22 & 71.40 \\
& TSV & 98.56 & 66.79 & \textbf{90.83} & 86.14 & 91.22 & 60.01 & 43.09 & 40.46 & 72.14 \\
& Iso-C & 98.51 & 66.89 & 90.90 & 86.20 & 90.32 & 57.52 & 33.31 & 34.55 & 69.78 \\
& Iso-CTS & 98.50 & 67.63 & 90.46 & 86.21 & 89.94 & 55.99 & 28.71 & 36.26 & 69.21 \\
& \textbf{SCORE (Ours)} & \textbf{98.62} & \textbf{68.26} & \underline{90.56} & \textbf{86.54} & \underline{91.28} & \textbf{61.19} & \underline{46.77} & \textbf{41.12} & \textbf{73.04} \\
& Task Experts & 99.92 & 81.62 & 91.58 & 88.10 & 96.15 & 88.61 & 67.88 & 60.84 & 84.34 \\
\bottomrule
\end{tabular}
\caption{Results for the leave-one-domain-out protocol: we measure accuracy (in \%) on the left-out domain of each column. On average, our SCORE outperforms all existing solutions. Columnwise, best results in \textbf{bold} and second-best \underline{underlined}: SCORE beats or stays very close to the best previous method.}
\label{tab:combined-results}
\end{table*}
We evaluate three CLIP~\cite{radford2021learning} variants (ViT-B/32, ViT-B/16, and ViT-L/14) on a comprehensive benchmark of eight domain generalization image classification datasets. The benchmark includes six natural-image datasets: PACS~\cite{pacs}, DomainNet~\cite{domainnet}, ImageNet-R~\cite{imagenetr}, NICO++~\cite{nicopp}, OfficeHome~\cite{officehome}, and TerraIncognita~\cite{terraincognita}; and two medical datasets, FedISIC~\cite{fedisic} and RetinaDomains~\cite{aptos2019,messidor,idriddataset,ddrdataset}.
In total, these datasets cover 49 different domains with varying label spaces, ranging from 4 to 365 classes, and capture a broad spectrum of distribution shifts. Per-domain results for each dataset appear in the Appendix~\ref{supp:per-domain-results}. 

We compare our approach with the following model merging methods: Task Arithmetic~\cite{ilharcoediting2023}, TIES~\cite{yadav2024ties}, MagMax~\cite{marczak2024magmax}, PCB~\cite{pcb}, TSV~\cite{tsv2025}, ISO-C~\cite{isoc2025}, and ISO-CTS~\cite{isoc2025}. We report the  absolute accuracy for natural-images datasets, and balanced accuracy for medical datasets, since they are highly imbalanced. For reference, we include each model's zero-shot performance as a lower~bound, and tasks experts as upper-bound. The code to reproduce our work is available at ~\url{https://github.com/VirtualSpaceman/score_cvpr26}.

\myparagraph{Evaluation protocol.} We adopt the leave-one-domain-out evaluation protocol~\cite{gulrajanisearch2021,zhou2022domain}. For each held-out target domain, we merge the fine-tuned models from all remaining source domains and evaluate the merged model on the unseen target domain, without any additional training or adaptation. We repeat this process for every domain and report the mean performance across all target domains.

\myparagraph{Fine-tuning details.} We fully fine-tune the CLIP's image encoder with a batch size of $128$, a learning rate of \text{$1e$-$5$} coupled with a cosine annealing schedule, and AdamW optimizer~\cite{adamw} with weight decay $0.1$ while keeping the text encoder unchanged. We use the CLIP's text encoder as the final classification layer output and keep it frozen during fine-tuning. This fine-tuning recipe preserves the model's open-vocabulary nature without compromising accuracy~\cite{ilharcoediting2023,marczak2024magmax,tsv2025}.

\subsection{Domain generalization performance}

Table~\ref{tab:combined-results} compares the performance across several model merging methods for ViT-B-32 and ViT-L-14. Results for ViT-B-16 appear in the appendix. Our approach achieves the highest average accuracy across all methods, surpassing the next-best competitor by 0.74 percentage points (p.p.) on the ViT-B-32 model and 0.58 p.p.~on the ViT-L-14 model. For the ViT-B-32, our method outperforms all competitors on the DomainNet dataset by 0.15 p.p., on NICOpp by 0.40~p.p., and on OfficeHome by 0.49 p.p. The performance gains are even more pronounced with the larger ViT-L-14 model. Our approach achieves the best results on PACS, tying for first place with a 0.02 p.p.~advantage over the second-best, and on DomainNet, where we are also tied for first, exceeding the second-best by 0.63 p.p. Additionally, we outperform competitors by 1.01 p.p.~on NICOpp and by 1.18 p.p.~on TerraIncognita, while remaining highly competitive on ImageNetR and OfficeHome.

When evaluating on specialized medical datasets, all methods show performance variations with model scale. With the ViT-B/32 model, our approach ranks second on the FedISIC dataset. Using the larger ViT-L/14 model, it again secures the second-best result on FedISIC and achieves the top performance on RetinaDomains, demonstrating strong adaptability across challenging medical imaging domains.

\subsection{Analysis and ablations}
\label{subsec:analysis-ablations}

\paragraph{Model merging can outperform zero-shot evaluation on unseen domains.} We investigate whether model merging can serve as a compositional mechanism to integrate knowledge from models fine-tuned on different distribution shifts. Specifically, we ask whether any merging strategy can surpass the zero-shot baseline in out-of-distribution generalization. Intuitively, combining models trained on complementary shifts should yield a merged model that generalizes better to unseen domains. However, this outcome is not guaranteed. If the compositional knowledge captured by the merged weights does not align with features that promote generalization, the merging process may instead degrade performance. 
% — altering parameters in directions that harm rather than enhance out-of-distribution robustness. 
Our study empirically tests this hypothesis, examining if model merging succeeds to outperform zero-shot evaluation in unseen environments. 

Table~\ref{tab:combined-results} indicates that most strategies outperform the zero-shot evaluation on out-of-distribution, with Task Arithmetic being the lone exception. The latter was one of the first strategies beyond the simple average that relies on delta weights, but it is sensitive to the choice of hyperparameter. Beyond these methods, recent methods, including ours, significantly outperform this baseline. These results support our initial hypothesis that practitioners can rely on model merging to combine multiple sources of information into a single model with improved generalization performance.  

These results indicate that model merging provides a practical and efficient strategy for combining multiple fine-tuned models to improve generalization under distribution shifts. This capability is valuable in dynamic or data-limited environments, where practitioners have access only to a fine-tuned model for certain distribution shifts, available through model repositories.  Performance gains are particularly evident on medical datasets. ll model merging strategies consistently improved over the zeroshot baseline across all model sizes by at least 6 p.p.~(FedISIC) and 8 p.p.~(RetinaDomains) for ViT-B-32, comparing the baseline with the worst model merging strategy on that particular dataset.

\myparagraph{Model merging outperforms model ensembles in domain generalization.} We compare model merging against traditional model ensembling to understand their relative effectiveness and efficiency. Model merging operates directly in parameter space, producing a single merged model that carries the combined knowledge of multiple fine-tuned models without increasing inference cost. In contrast, model ensembling combines predictions in output space, which scales poorly as the number of source domains increases. Each additional model in an ensemble adds memory overhead, computational cost, and inference latency, since every model must be loaded in memory, and its outputs aggregated at test time. 

Table~\ref{tab:modelmerging-ensemble} compares the best-performing model-merging method 
according to our experiments with the logit ensemble, which is a common ensemble strategy~\cite{ganaie2022ensemble,yang2023survey} on the same 8 domain generalization datasets. Our proposed SCORE method consistently outperforms the logit ensemble baseline by 1.12 to 1.90 p.p.~across all model sizes. Critically, model merging maintains the inference cost of a single model, thereby alleviating the computational and memory overhead of an ensemble's multiple forward passes. Surpassing the ensemble's performance, rather than merely matching it, demonstrates that SCORE successfully integrates complementary knowledge from the fine-tuned \text{models.} 

\definecolor{posgreen}{RGB}{0,120,0}
\definecolor{negred}{RGB}{200,0,0}

\begin{table}[t]
\centering
\scriptsize

\caption{Mean average accuracy (\%) across datasets for different ViT architectures.
Model ensemble is the logit ensemble baseline; values in parentheses denote
relative changes. Improvements are shown in \textcolor{posgreen}{green}.}
\setlength{\tabcolsep}{3.5pt}
\renewcommand{\arraystretch}{1.05}
\begin{tabular}{lccc}
\toprule
\textbf{Method} & \textbf{ViT-B-32} & \textbf{ViT-B-16} & \textbf{ViT-L-14} \\
\midrule
Model ensemble & \underline{64.57}  & \underline{68.07} & \underline{71.81}  \\
\textbf{SCORE (Ours)} & \textbf{65.69} {(\textcolor{posgreen}{+1.12})} &
\textbf{69.97} {(\textcolor{posgreen}{+1.90})} &
\textbf{73.04} {(\textcolor{posgreen}{+1.24})} \\
\bottomrule
\end{tabular}

\label{tab:modelmerging-ensemble}
\end{table}

Our findings expand the previous evaluation by \citet{wortsman2022model}, which considered only weight averaging and distribution shifts on ImageNet. In this study, we consider a broader range of datasets for domain generalization and model-merging candidates, using a typical evaluation scheme from the domain generalization literature.

\myparagraph{The impact of the trim function on performance.}
SCORE employs a \textit{trim} operation to mitigate conflicts among singular directions and enhance generalization. It retains the main diagonal while removing off-diagonal outliers from $\Sigma_{\text{score}}$ to reduce conflicting directions. Table~\ref{tab:ablations-sigma} shows
our ablation study to considering different strategies for $\Sigma_{\text{score}}$:

\begin{itemize}
    \item \textbf{Diagonal only:} keeping only the main diagonal elements;
    \item \textbf{Off-diagonal only:} considering only the off-diagonal components (keep only the domain-wise conflicts);
    \item \textbf{Full matrix:} preserving both diagonal and off-diagonal elements;
    \item \textbf{Trimmed:} retaining both diagonal and off-diagonal elements but trimming the off-diagonal outliers ($trim$ operation). Our proposed SCORE relies on this strategy. 
\end{itemize}

\definecolor{posgreen}{RGB}{0,120,0}
\definecolor{negred}{RGB}{200,0,0}

\begin{table}[htb!]
\centering
\scriptsize
\caption{Ablation on the choice of merged task matrix. We report
the mean average accuracy (\%) across domains for each ViT architecture and, in parenthesis, the \textcolor{posgreen}{positive} or \textcolor{negred}{negative} difference to the baseline given by Diagonal. The proposed trimmed merging offers the best results.}
\begin{tabular}{lccc}
\toprule
\textbf{Method} & \textbf{ViT-B-32} & \textbf{ViT-B-16} & \textbf{ViT-L-14} \\
\midrule
Diagonal
  & \underline{63.62} 
  & \underline{67.38} 
  & \underline{71.50} \\
\addlinespace[2pt]
Off-diagonal
  & 58.41 (\textcolor{negred}{-5.21})
  & 62.41 (\textcolor{negred}{-4.97})
  & 67.46 (\textcolor{negred}{-4.04}) \\
\addlinespace[2pt]
Full matrix
  & 7.59 (\textcolor{negred}{-56.03})
  & 7.66 (\textcolor{negred}{-59.72})
  & 7.70 (\textcolor{negred}{-63.80}) \\
\addlinespace[2pt]
\textbf{Trimmed}
  & \textbf{65.69} (\textcolor{posgreen}{+2.07})
  & \textbf{69.97} (\textcolor{posgreen}{+2.59})
  & \textbf{73.04} (\textcolor{posgreen}{+1.53}) \\
\bottomrule
\end{tabular}

\label{tab:ablations-sigma}
\end{table}

We use the configuration that keeps only the main diagonal elements of $\Sigma_{score}$ as our baseline. Retaining only the off-diagonal components reduces average performance across all architectures by up to 5 p.p., indicating that these components still carry meaningful information for generalization. However, preserving both diagonal and off-diagonal elements without trimming leads to a substantial performance drop, suggesting strong interference and competition between singular directions, promoting high feature competition in the merged model. In contrast, our proposed strategy via the $trim$ function balances the contributions of shared subspaces while removing destructive outliers, improving by up to 2.59 p.p.~in average performance relative to the diagonal-only baseline.  
\section{Conclusion}
\label{sec:conclusion}

We studied model merging as a compositional mechanism for integrating multiple fine-tuned models trained on distinct domains and evaluated the merged models'  performance on unseen domains. We introduced SCORE, a novel model merging approach to mitigate conflicts over singular directions. Our method outperforms, on average, all other model merging methods for 8 domain-generalization datasets and 3 model sizes. Also, SCORE model-merging outperforms a model-ensembling baseline. Our main contribution is agnostic to the specifics of the domain-generalization method and depends only on practitioners fine-tuning from a common base model. For instance, future research could explore the advantages of combining multiple models, each fine-tuned on the same multi-source domain data but using different domain-generalization losses. Our work also opens a new avenue for using model merging as a strategy to investigate generalization capabilities, particularly its compositional abilities, as in ~\cite{mayilvahanandoes2024,kempfand2025}. For instance, instead of retraining models from scratch on several domain subsets, one could merge individual models and examine whether the compositional results of~\citet{kempfand2025} still hold.

\myparagraph{Limitations.} Our experimental design assumes we have access only to the fine-tuned model for a given distribution shift, not to the source data. 
A limitation of every parameter-wise model merging is only merging models that share the same architecture and fine-tune from the same seed model. While our work primarily focuses on image classification, we anticipate future extensions to natural language processing and generative models, including large language models and image generative models.

\myparagraph{Acknowledgements} LC is partially funded by FAPESP (2024/16685-7), CAPES, and Becas Santander/UNICAMP 2022. SA is also partially funded by FAPESP (2023/12086-9, 2023/12865-8, 2020/09838-0, 2013/08293-7), H.IAAC 01245.003479/2024-10 and CNPq 316489/2023-9. CZ and RB acknowledge the Gauss Centre for Supercomputing e.V.~for funding this project by providing computing time on the~GCS Supercomputer JUWELS at Jülich Supercomputing Centre (JSC). They also acknowledge funding from the European Research Council~(ERC) under the Horizon Europe Framework Programme (HORIZON) for proposal number 101116395 \text{SPARSE-ML.}
{
    \small
    \bibliographystyle{ieeenat_fullname}
    \bibliography{main}
}

\clearpage
\setcounter{page}{1}
\maketitlesupplementary
\renewcommand{\thesection}{A\arabic{section}}
\setcounter{section}{0}

\section{Dataset details}
\label{supp:dataset_details}
Here, we describe the datasets we use in this worn and include the number of domains and classes for each dataset. 

\begin{itemize}
    \item \textbf{PACS~\cite{pacs}} is a small-scale domain-generalization benchmark consisting of four distinct visual styles (domains): Photo, Art painting, Cartoon and Sketch. It contains 7 object classes (dog, elephant, giraffe, guitar, horse, house, person) and about 9,991 images. 
    \item \textbf{DomainNet~\cite{domainnet}} is a large-scale multi-domain dataset designed for multi-source  generalization. It contains 345 object categories spanning six very different domains (clipart, infograph, painting, quickdraw, real (photo), and sketch), with roughly 586k images. 
    
    \item \textbf{ImageNetR~\cite{imagenetr}} is a rendition-style robustness built from ImageNet labels: renditions such as paintings, cartoons, graffiti, sculptures, toys, etc. It covers 200 ImageNet classes and contains on the order of 30k images. 
    \item \textbf{NICOpp~\cite{nicopp}} designed to challenge models that learn spurious correlations by grouping images into ``contexts'' (\eg, ``dog on a beach'') rather than just broad styles.
    \item \textbf{OfficeHome~\cite{officehome}} it consists of images from 4 different domains: Artistic images, Clip Art, Product images and Real-World images. For each domain, the dataset contains images of 65 object categories found typically in everyday office and home objects.
    \item \textbf{TerraIncognita~\cite{terraincognita}} it features camera trap images of 10 animal species from four different geographical locations, which serve as the domains. The main objective is specie classification across these domains. 
    \item \textbf{FedISIC} comprises dermoscopy images collected from four hospitals. We use same splits of~\citet{fedisic} containing 23,247 images from the public training of ISIC2019~\cite{tschandl2018ham10000,hernandez2024bcn20000,codella2019skin}. The task involves classifying images into eight melanoma classes, which presents a significant label imbalance. The domains are based on the imaging acquisition device. We have 6 distinct domains.
    \item \textbf{RetinaDomains} composed of a collection of four datasets: Aptos-2019~\cite{aptos2019}, Messidor~\cite{messidor}, IDRDI~\cite{idriddataset}, DDR~\cite{ddrdataset}. The classification task is five Diabetic Retinopathy grading, in which each class corresponds to a grading. We use each dataset as a single domain. 
\end{itemize}

\section{Experimental details}
\label{supp:expeerimental_details}
We provide additional details regarding the implementation details and the additional per-domain performances across all datasets and model architectures considered in this work. 

\subsection{Implementation details.}
\label{supp:implementation_details}
Our method relies on SVD decomposition, which is defined for any two-dimensional matrix $\Delta~w~\in~\mathbb{R}^{m~\times~n}$. However, in the model architecture we consider one-dimensional weights also exists. In this case, we follow previous works of ~\citet{tsv2025,isoc2025} and take the average over all parameters. 

We compute the per-matrix $\sigma_{\text{off}}$ as shown in Equation~\ref{eq-trim}. Since medical datasets are particularly sensitive to off-diagonal noise, we also normalize the $\sigma_{\text{off}}$ by the number of merged domains.

\subsection{Fine-tuning details.}
We fully fine-tune the CLIP's image encoder with a batch size of $128$, a learning rate of \text{$1e$-$5$} coupled with a cosine annealing schedule, and AdamW optimizer~\cite{adamw} with weight decay $0.1$ while keeping the text encoder unchanged. We fine-tune, for each domain, a CLIP's visual encoder.  We adopt 20 epochs for FedISIC, and RetinaDomains (medical datasets), NICOpp for 15 epochs and 10 epochs otherwise. 

\subsection{Multi-task learning performance}
\label{supp:subsec-mtl-performance}
We compare SCORE to three methods: Task Arithmetic, TIES, and TSV. The latter is the most cited peer-reviewed method from 2025 in our scorer's suite. We follow the Multi-Task Learning (MTL) scheme across 8 tasks from prior work, using ViT-B-32, ViT-B-16, and ViT-L-14 models. Average accuracy (in \%) is reported below. Our method \textbf{achieves high IID performance}, while enhancing OOD performance.

\begin{table}[!ht]
    \centering
    \scriptsize
    \caption{Average accuracy (in \%) of the merged model across 8 MTL datasets across 3 ViT models. Higher is better.}
    % \vspace{-0.3cm}
    \resizebox{0.70\columnwidth}{!}{%
    \begin{tabular}{lccc}
        \hline
        & ViT-B-32 & ViT-B-16 & ViT-L-14 \\
        \hline
        Task Arithm. & 69.98 & 75.38 & 82.91 \\ 
        TIES & 73.73 & 78.76 & 84.14 \\ 
        TSV & 83.70 & 87.23 & 90.53 \\ 
        \textbf{SCORE (Ours)} & \textbf{84.06} & \textbf{87.60} & \textbf{91.17} \\ 
        \hline
    \end{tabular}%
    }
    \label{tab:reb:mtl-performance}
\end{table}

\subsection{Other measures of subspace overlap} 
\label{supp:subsec-overlap}
\begin{figure}[htb!]
  \centering
  
  % First subfigure
  \begin{subfigure}[t]{0.47\columnwidth}
    \centering
    \includegraphics[width=\linewidth]{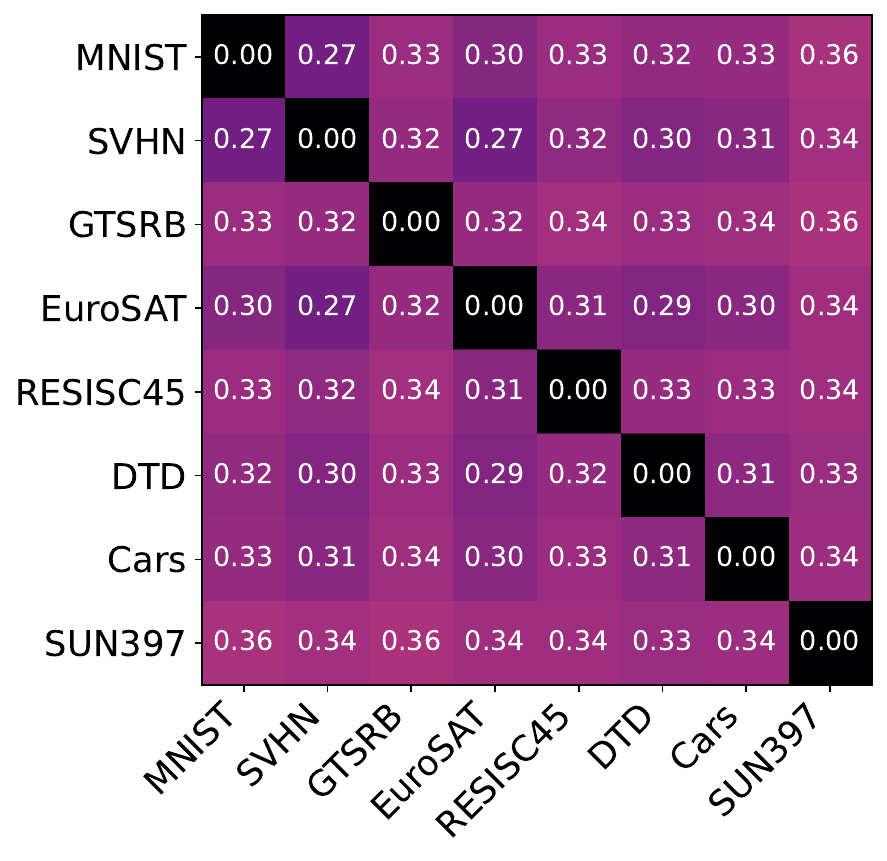}
    % \caption{Multi-task learning}
  \end{subfigure}
  \hfill
  % Second subfigure
  \begin{subfigure}[t]{0.48\columnwidth}
    \centering
    \includegraphics[width=\linewidth]{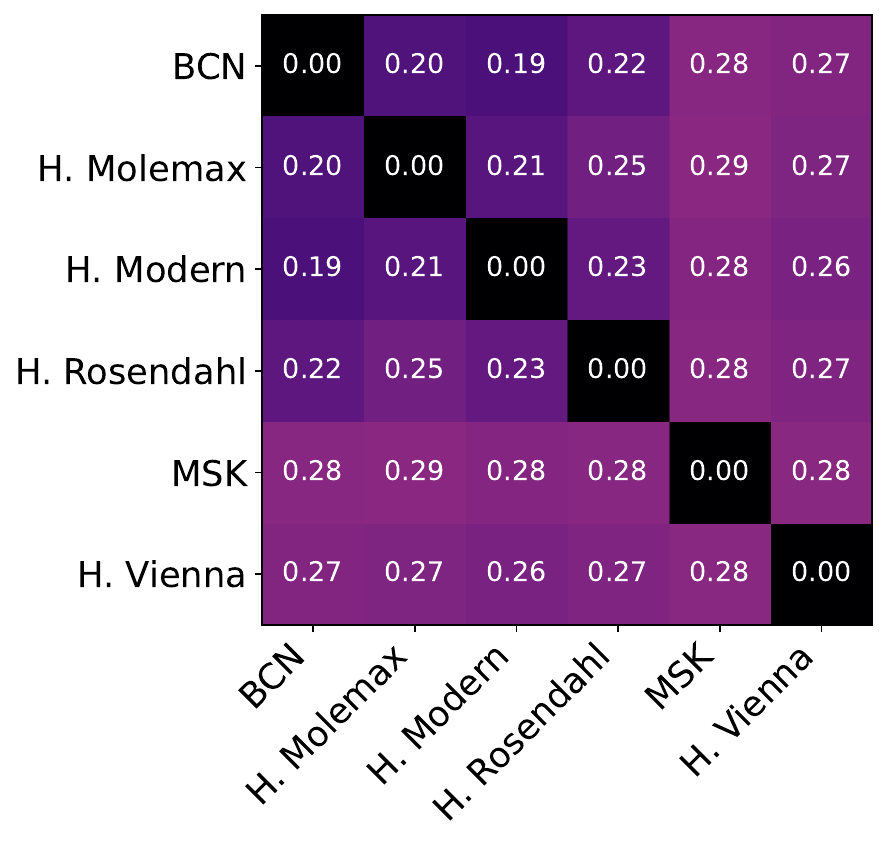}

  \end{subfigure}
  
  % Main caption for both figures
  \vspace{-0.5cm}
  \caption{Pairwise $\theta_{avg}$ (in radians). Models fine-tuned for 8~MTL datasets(left). Models fine-tuned for 6 domains of FedISIC (right). 
  % Lower values indicate more overlap.
  }
  \label{fig:mean-theta}
\end{figure}

We compute per-layer principal angles $\theta\in[0, \pi/2]$ between subspaces as $\theta = arccos(S)$, where $ USV^{T} =  SVD(A^{T}B)$, where $A, B\in\mathbb{R}^{m\times~d}$~\cite{subspace-method} are subspaces and report the average value in Figure~\ref{fig:mean-theta}. Still, DG presents higher overlap (lower values) measures than MTL.

\subsection{Per-domain results}
\label{supp:per-domain-results}
\begin{figure*}[htb!]
    \centering
    \includegraphics[width=\linewidth]{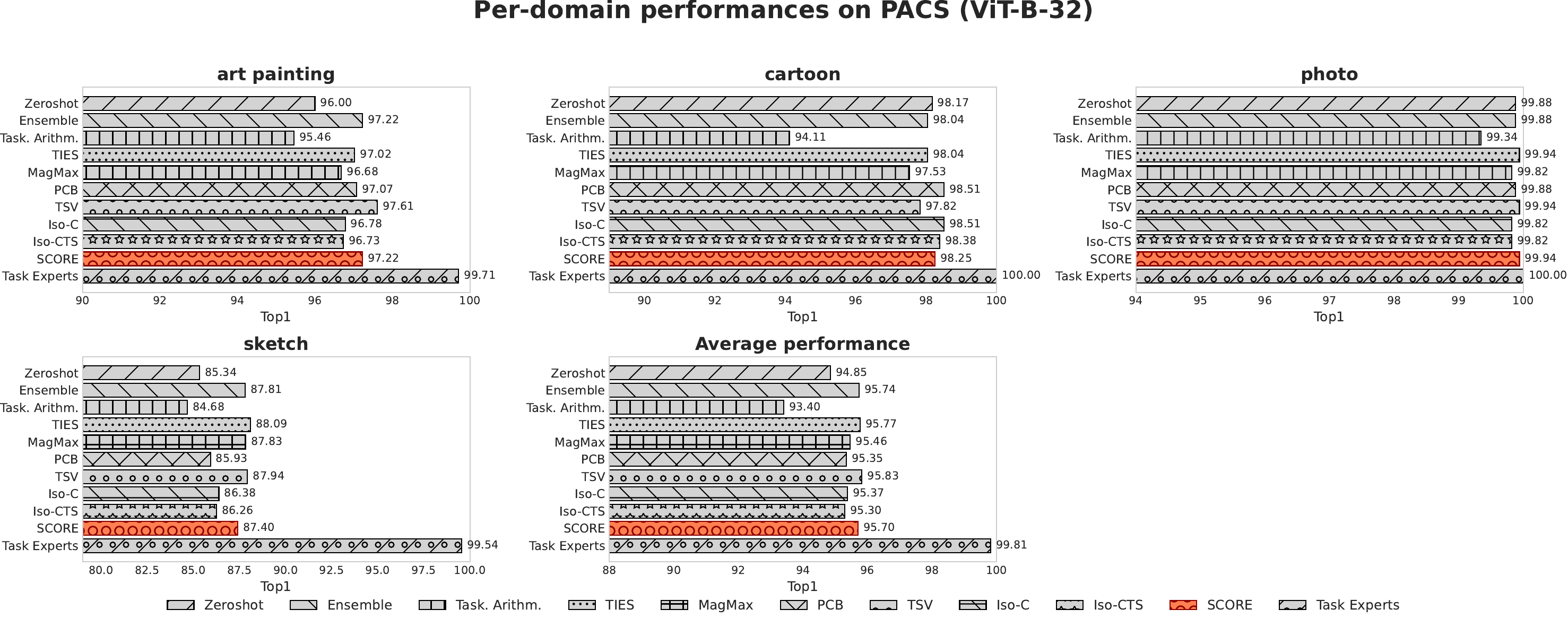}
    \caption{Per-domain results for ViT-B-32 on the PACS dataset for each model merging method in our study. }
    \label{fig:vitb32-pacs}
\end{figure*}

\begin{figure*}[htb!]
    \centering
    \includegraphics[width=\linewidth]{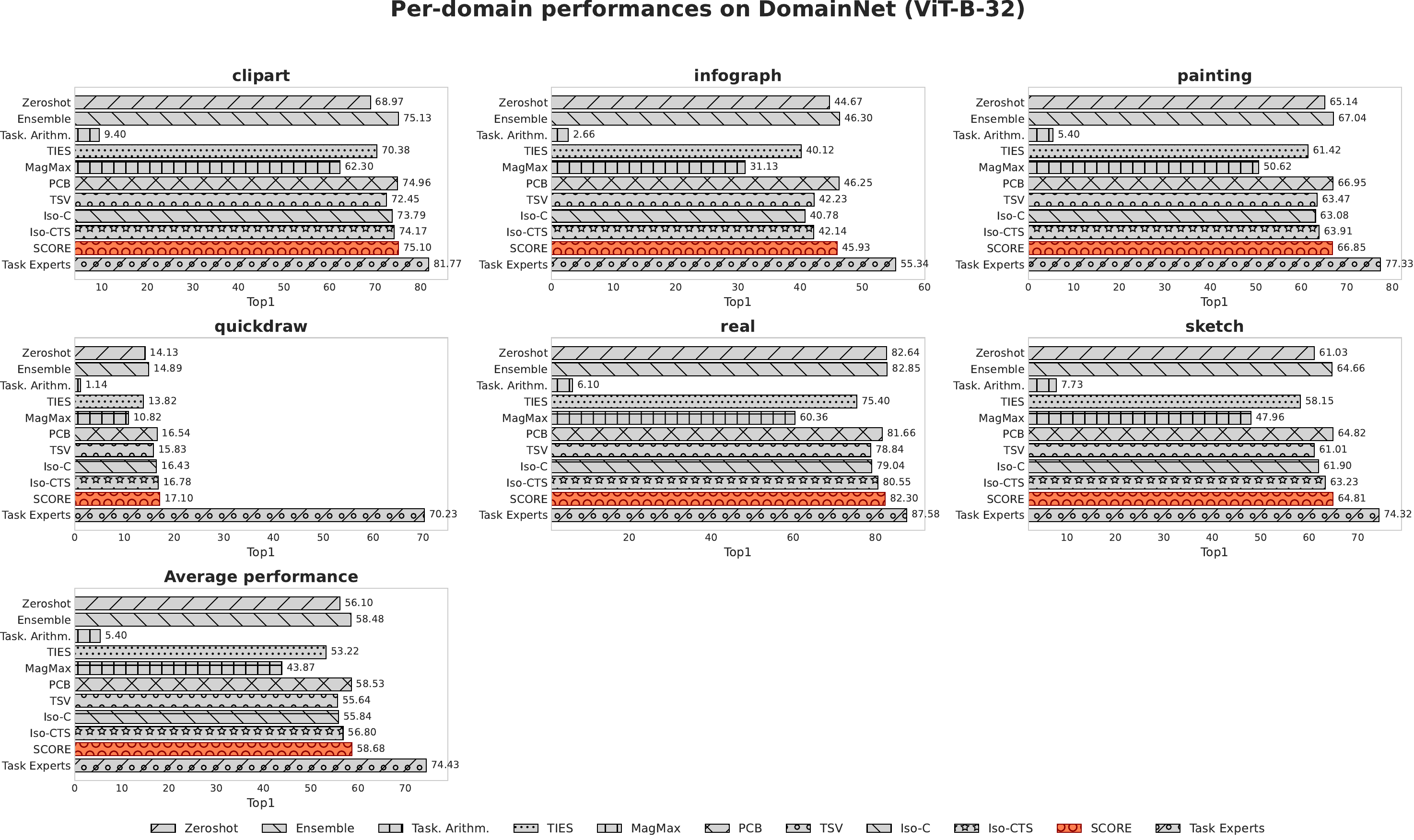}
    \caption{Per-domain results for ViT-B-32 on the DomainNet dataset for each model merging method in our study.}
    \label{fig:vitb32-domainet}
\end{figure*}

\begin{figure*}[htb!]
    \centering
    \includegraphics[width=\linewidth]{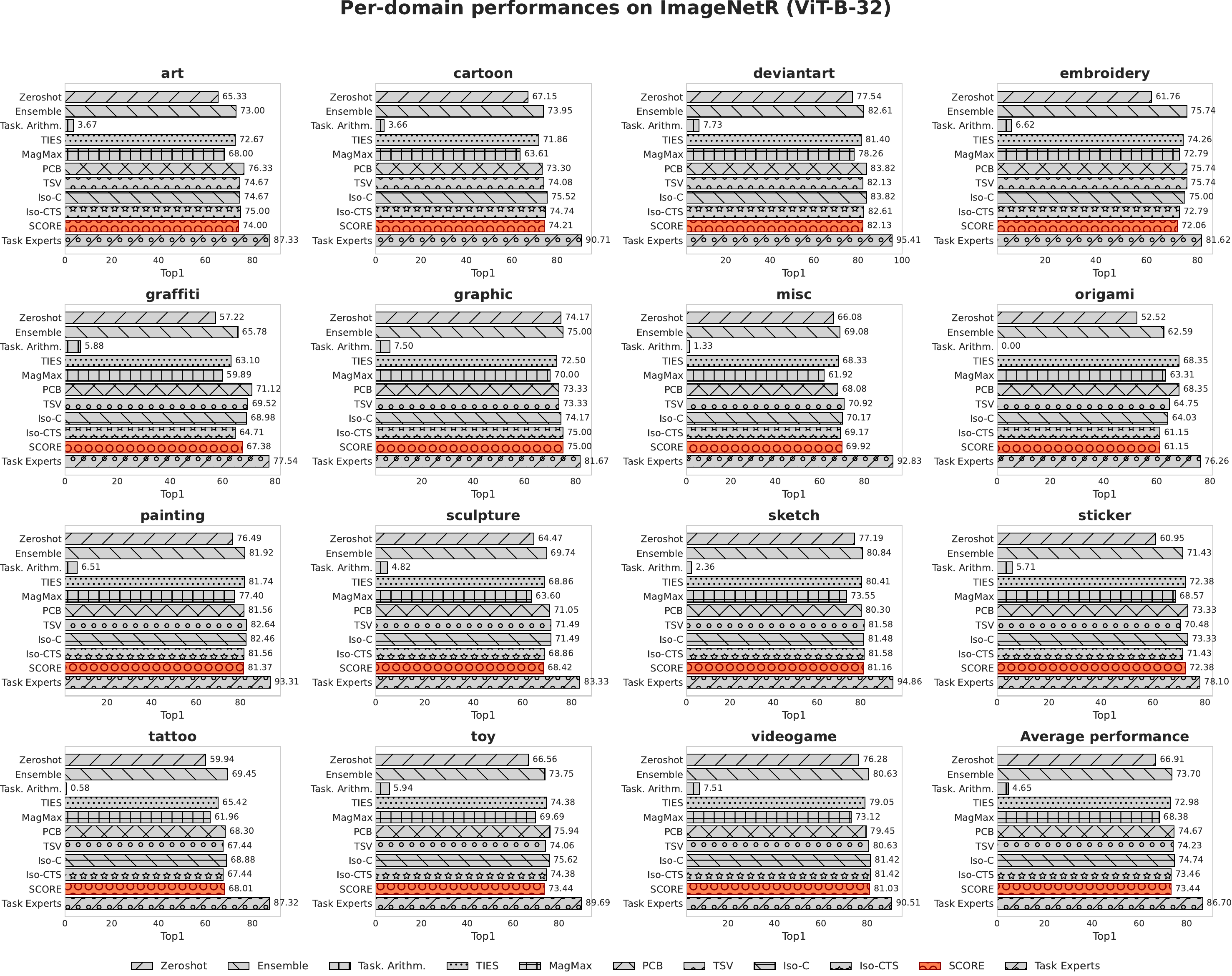}
    \caption{Per-domain results for ViT-B-32 on the ImageNetR dataset for each model merging method in our study.}
    \label{fig:vitb32-imagenetr}
\end{figure*}

\begin{figure*}[htb!]
    \centering
    \includegraphics[width=\linewidth]{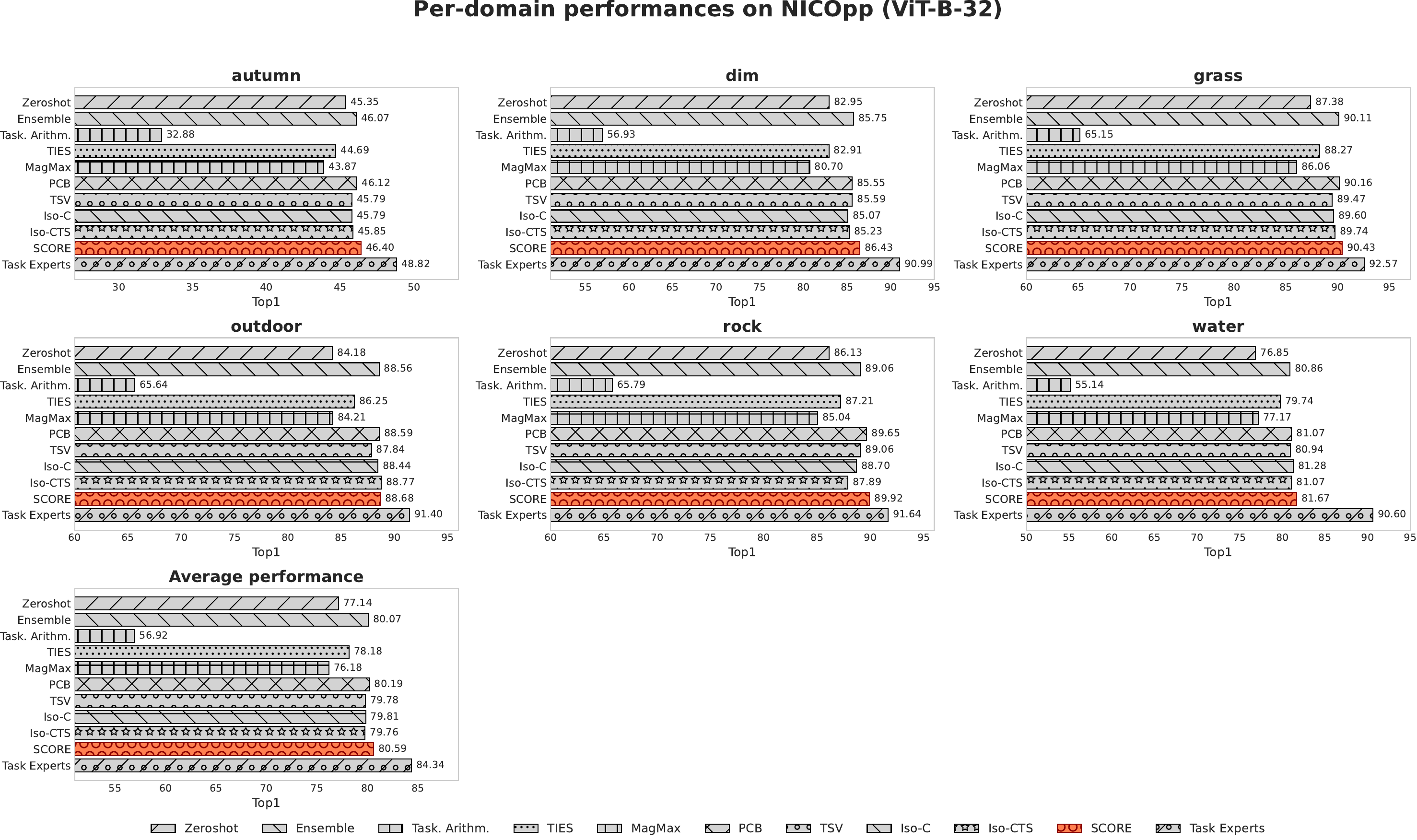}
    \caption{Per-domain results for ViT-B-32 on the NICO++ dataset for each model merging method in our study.}
    \label{fig:vitb32-nicopp}
\end{figure*}

\begin{figure*}[htb!]
    \centering
    \includegraphics[width=\linewidth]{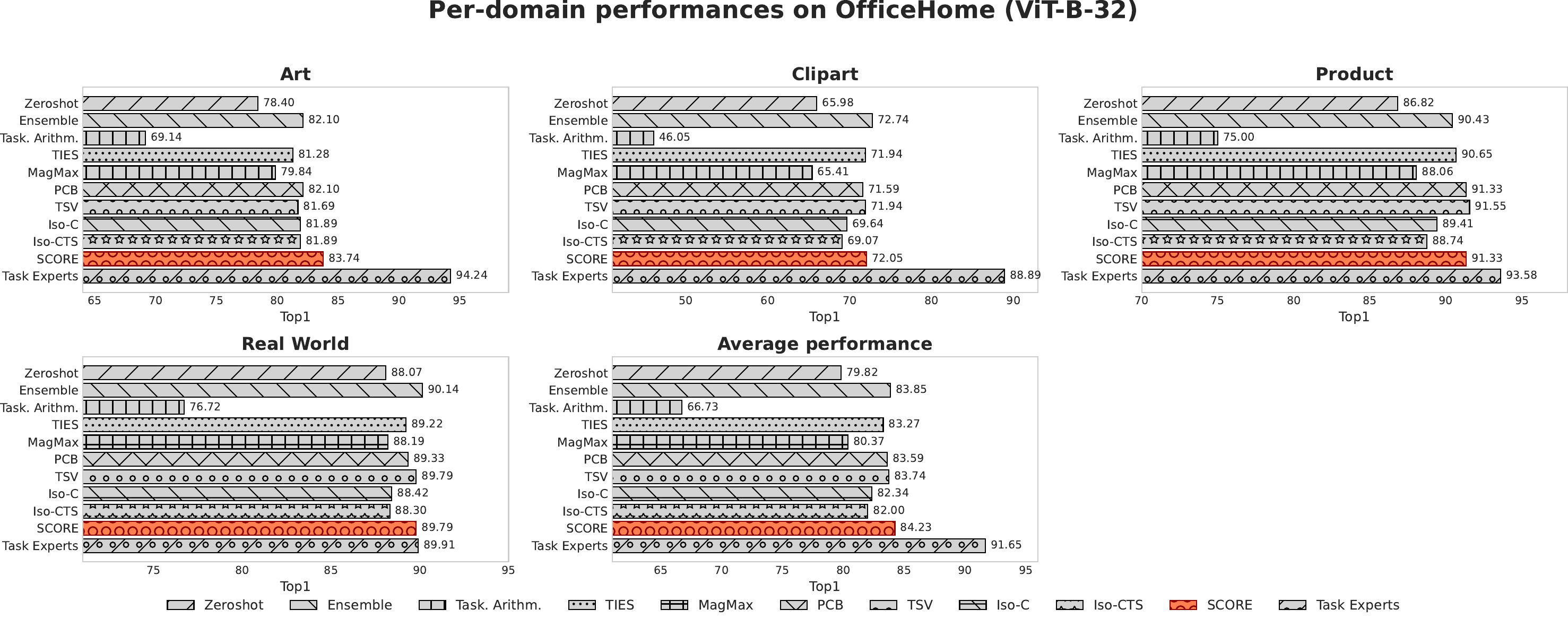}
    \caption{Per-domain results for ViT-B-32 on the OfficeHome dataset for each model merging method in our study.}
    \label{fig:vitb32-officehome}
\end{figure*}

\begin{figure*}[htb!]
    \centering
    \includegraphics[width=\linewidth]{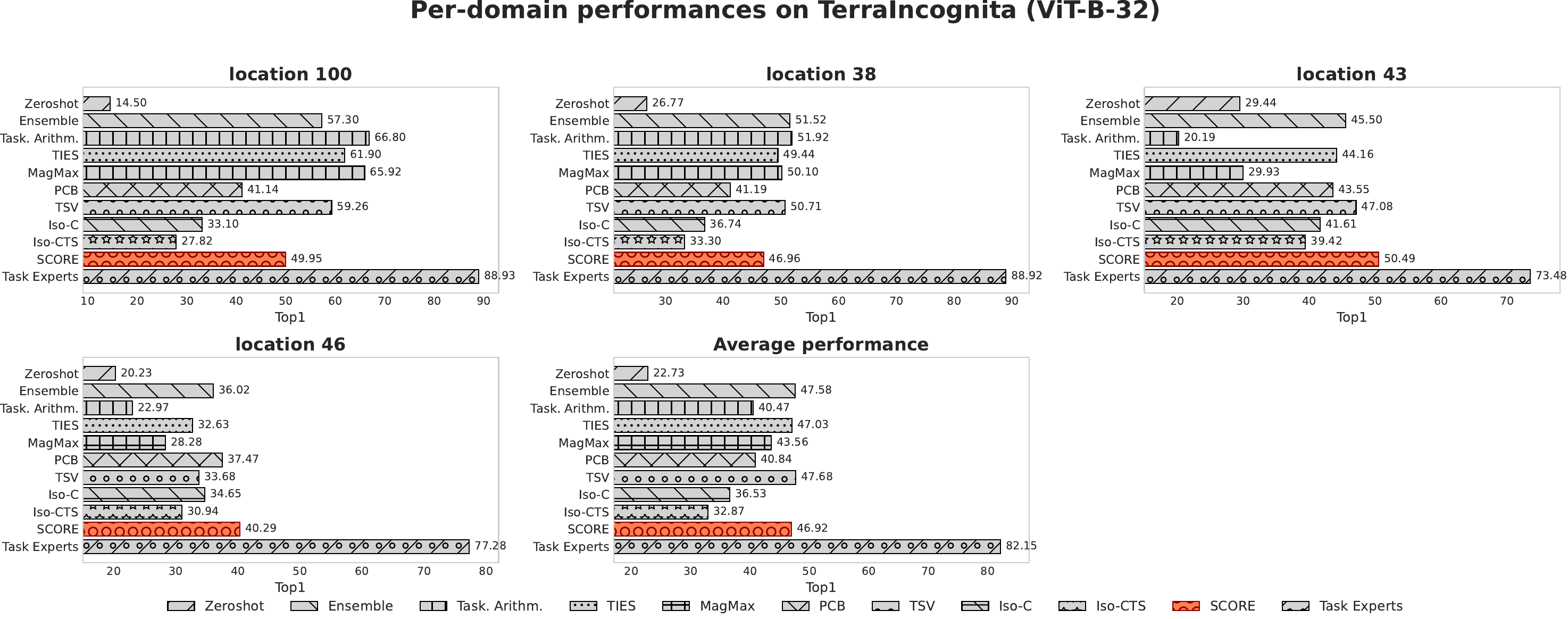}
    \caption{Per-domain results for ViT-B-32 on the TerraIncognita dataset for each model merging method in our study.}
    \label{fig:vitb32-terraincognita}
\end{figure*}

\begin{figure*}[htb!]
    \centering
    \includegraphics[width=\linewidth]{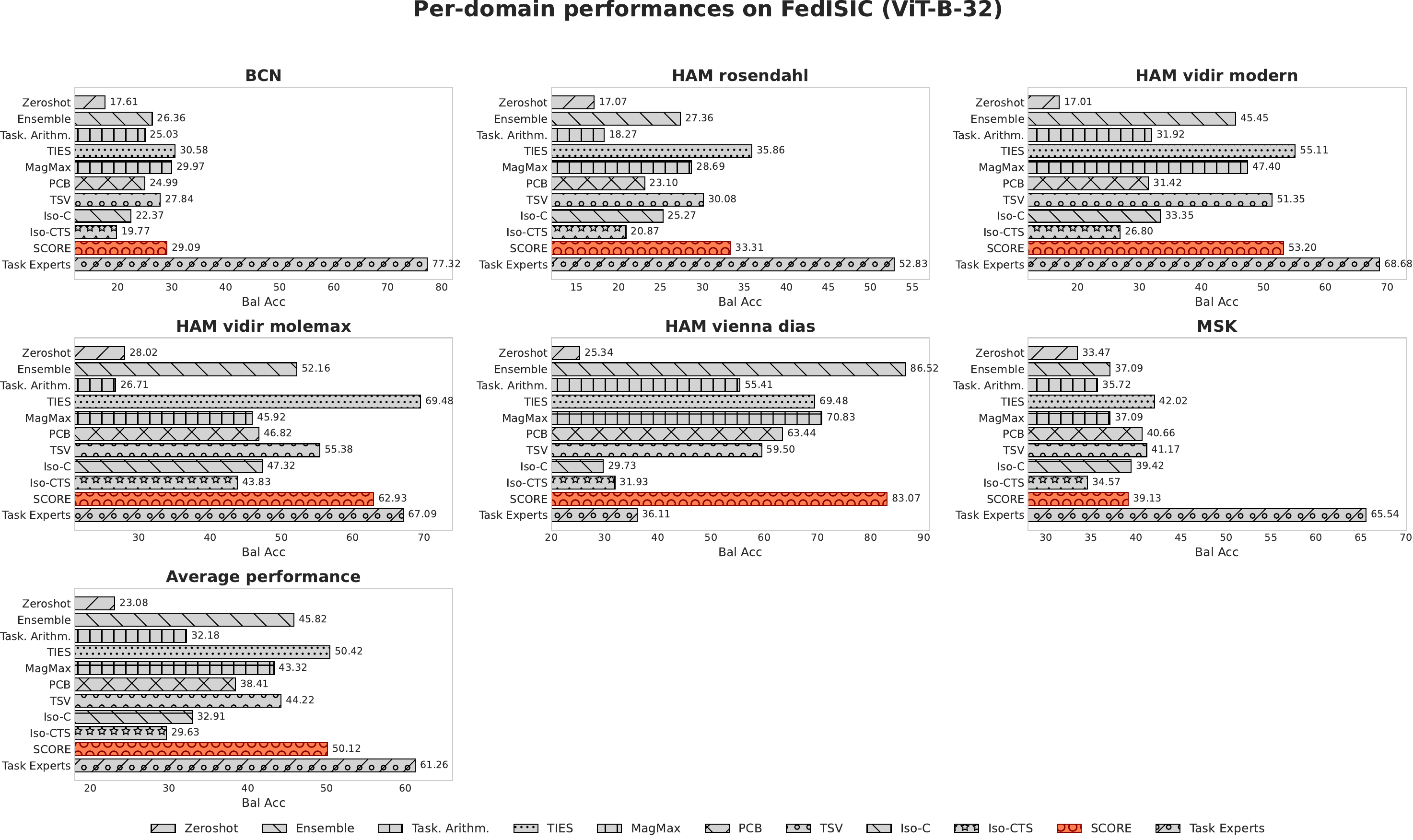}
    \caption{Per-domain results for ViT-B-32 on the FedISIC dataset for each model merging method in our study.}
    \label{fig:vitb32-fedisic}
\end{figure*}

\begin{figure*}[htb!]
    \centering
    \includegraphics[width=\linewidth]{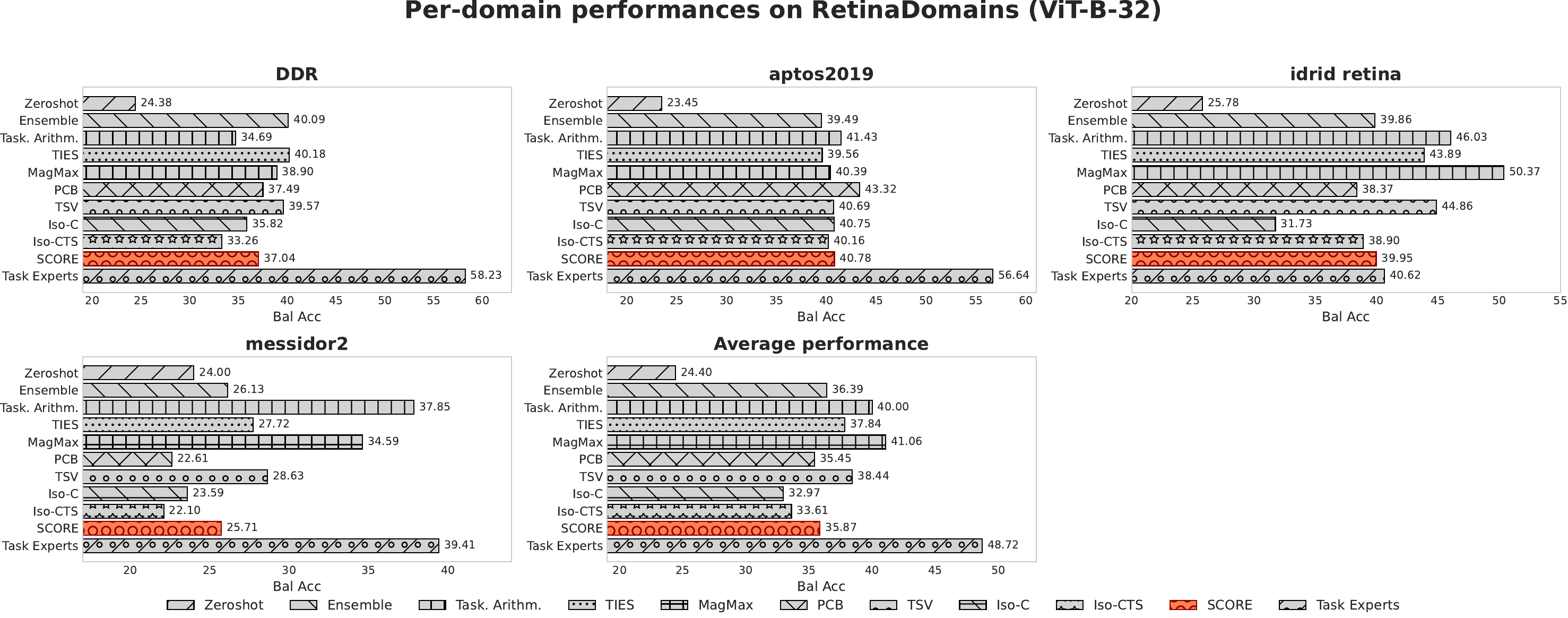}
    \caption{Per-domain results for ViT-B-32 on the RetinaDomains dataset for each model merging method in our study.}
    \label{fig:vitb32-retina}
\end{figure*}

\begin{figure*}[htb!]
    \centering
    \includegraphics[width=\linewidth]{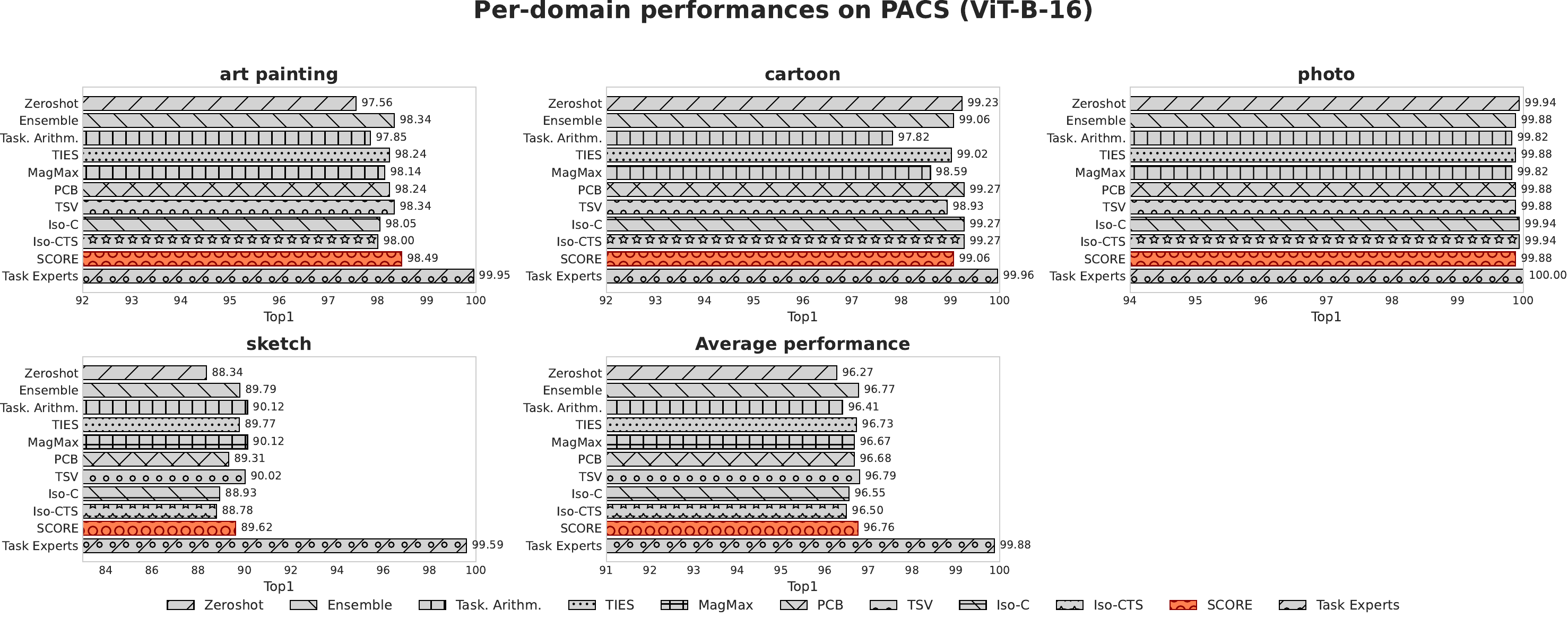}
    \caption{Per-domain results for ViT-B-16 on the PACS dataset for each model merging method in our study.}
    \label{fig:vitb16-pacs}
\end{figure*}

\begin{figure*}[htb!]
    \centering
    \includegraphics[width=\linewidth]{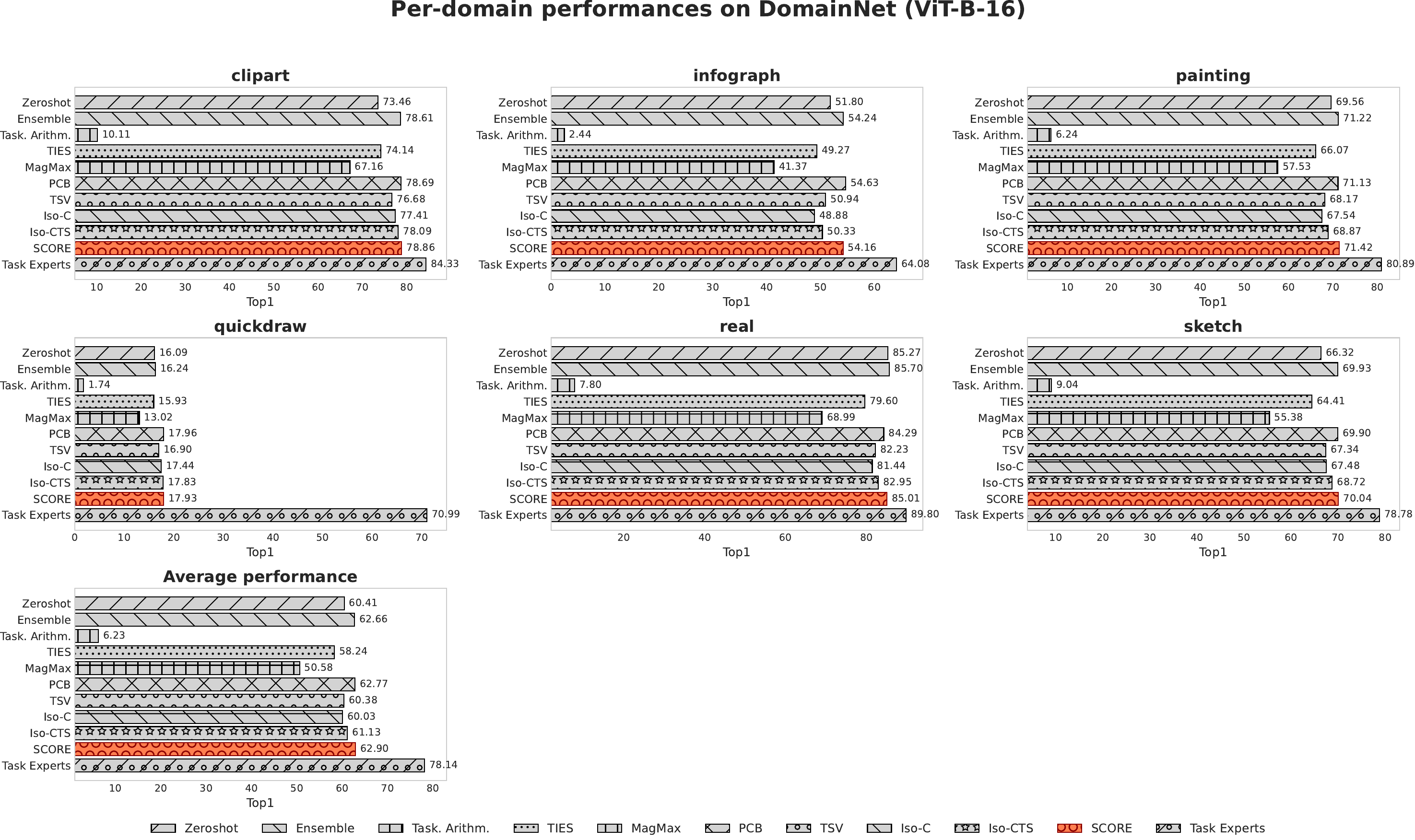}
    \caption{Per-domain results for ViT-B-16 on the DomainNet dataset for each model merging method in our study.}
    \label{fig:vitb16-domainet}
\end{figure*}

\begin{figure*}[htb!]
    \centering
    \includegraphics[width=\linewidth]{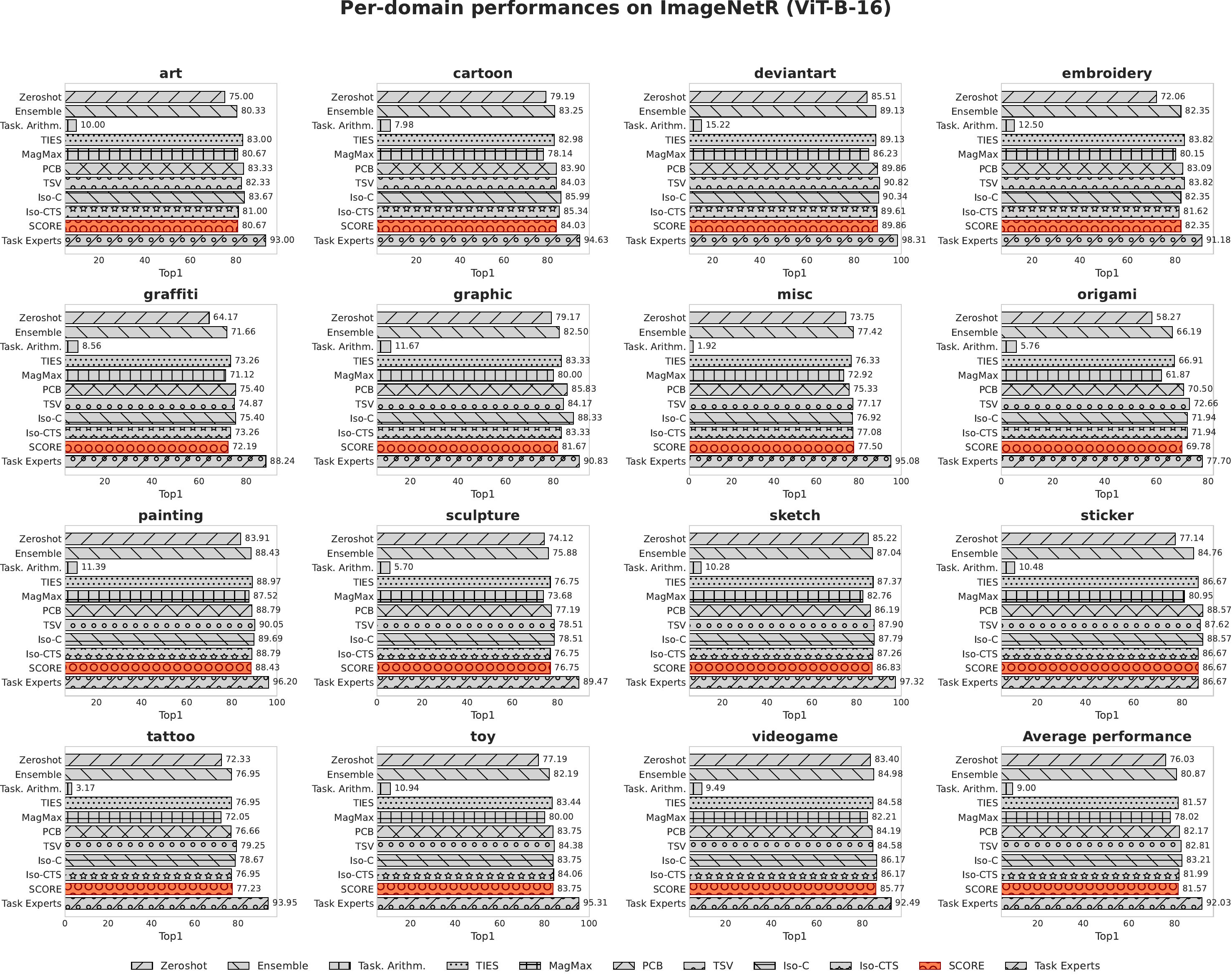}
    \caption{Per-domain results for ViT-B-16 on the ImageNetR dataset for each model merging method in our study.}
    \label{fig:vitb16-imagenetr}
\end{figure*}

\begin{figure*}[htb!]
    \centering
    \includegraphics[width=\linewidth]{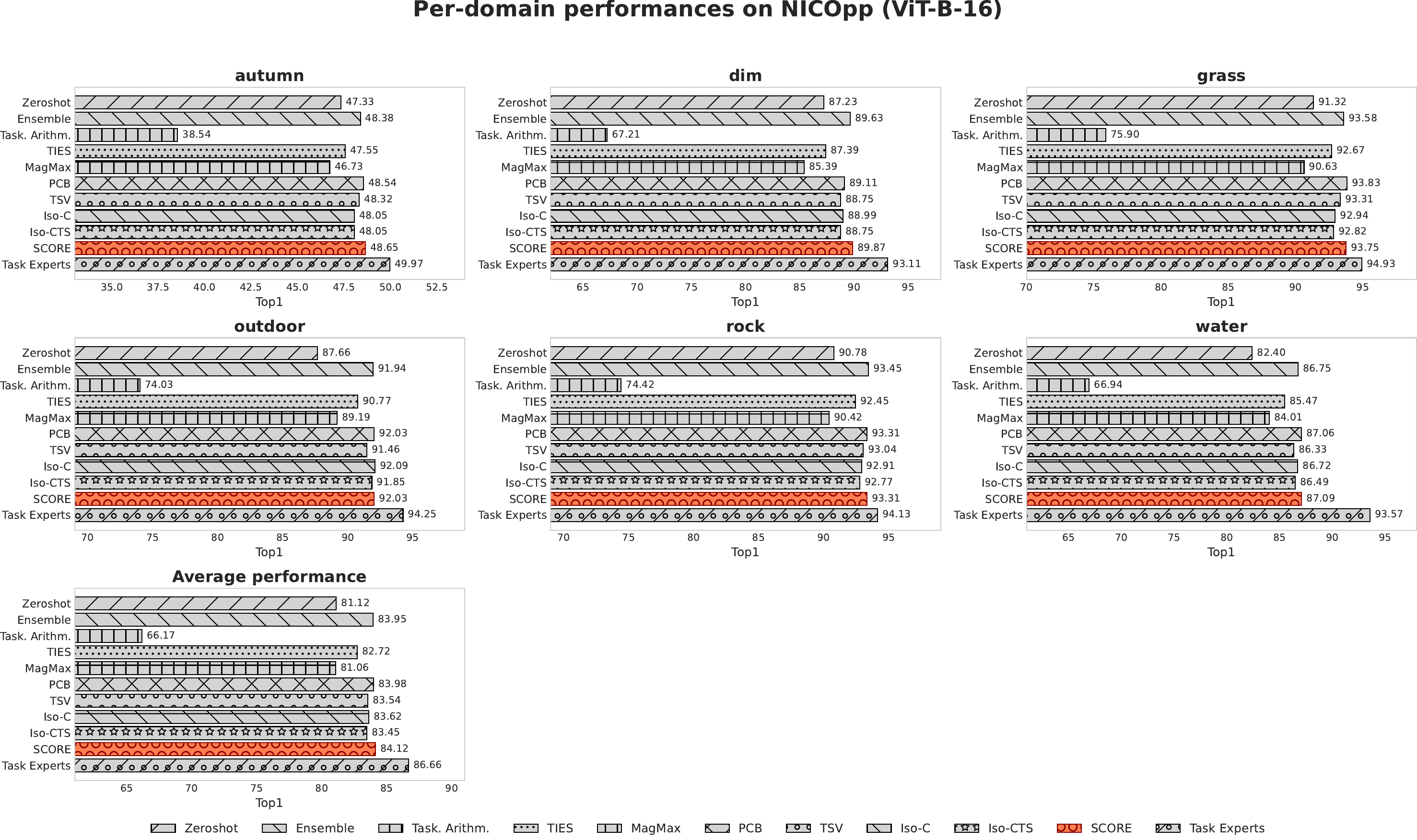}
    \caption{Per-domain results for ViT-B-16 on the NICO++ dataset for each model merging method in our study.}
    \label{fig:vitb16-nicopp}
\end{figure*}

\begin{figure*}[htb!]
    \centering
    \includegraphics[width=\linewidth]{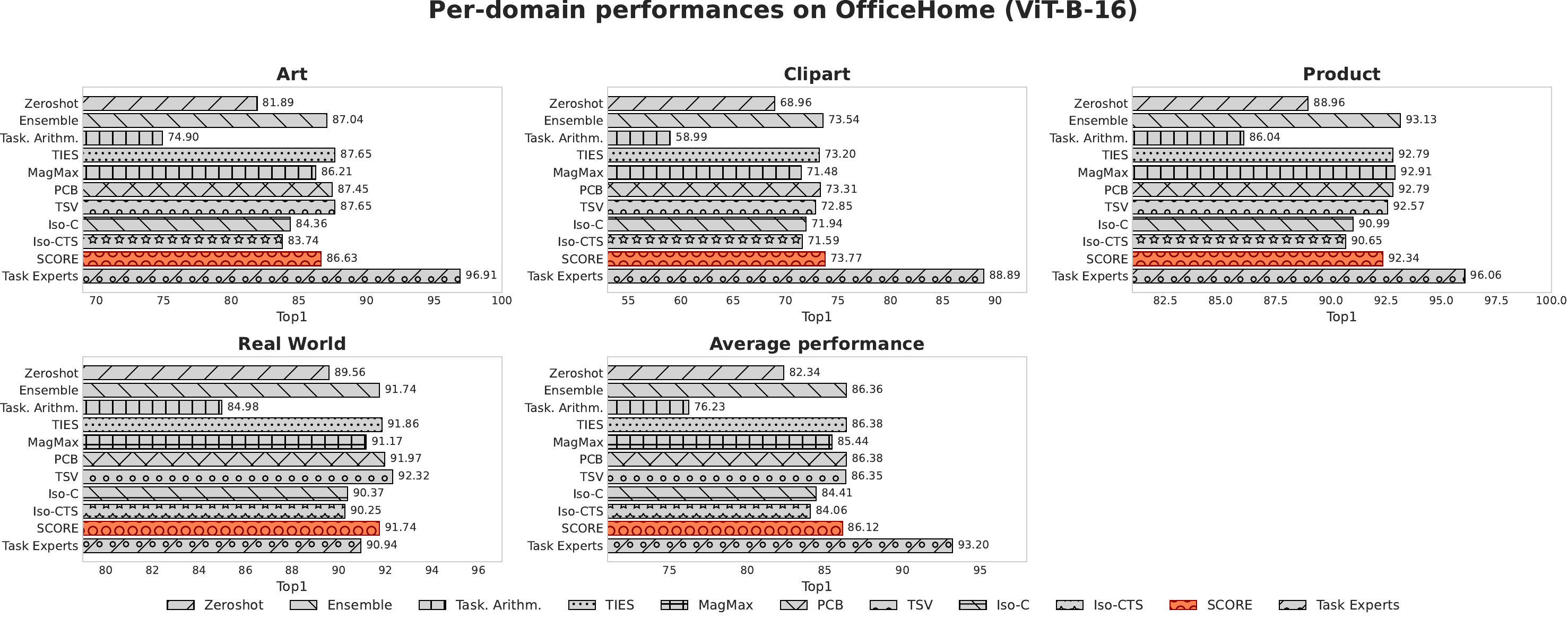}
    \caption{Per-domain results for ViT-B-16 on the OfficeHome dataset for each model merging method in our study.}
    \label{fig:vitb16-officehome}
\end{figure*}

\begin{figure*}[htb!]
    \centering
    \includegraphics[width=\linewidth]{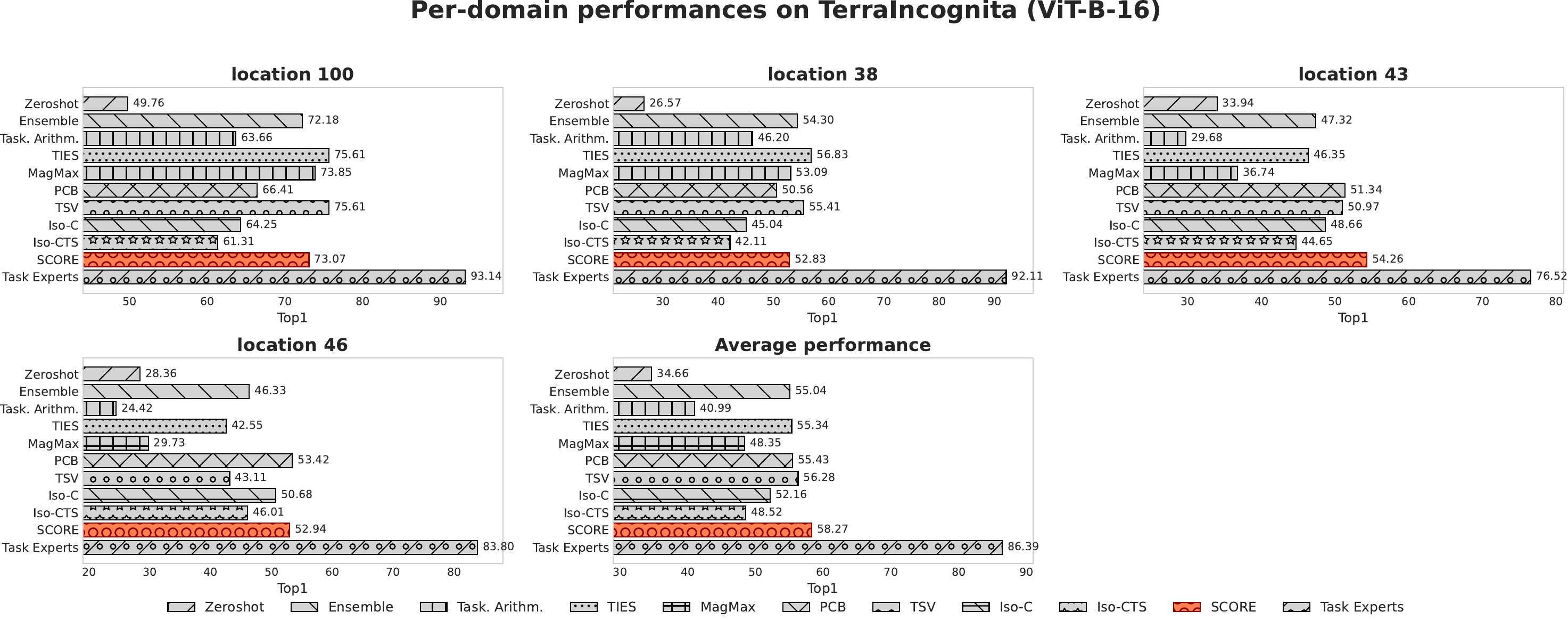}
    \caption{Per-domain results for ViT-B-16 on the TerraIncognita dataset for each model merging method in our study.}
    \label{fig:vitb16-terraincognita}
\end{figure*}

\begin{figure*}[htb!]
    \centering
    \includegraphics[width=\linewidth]{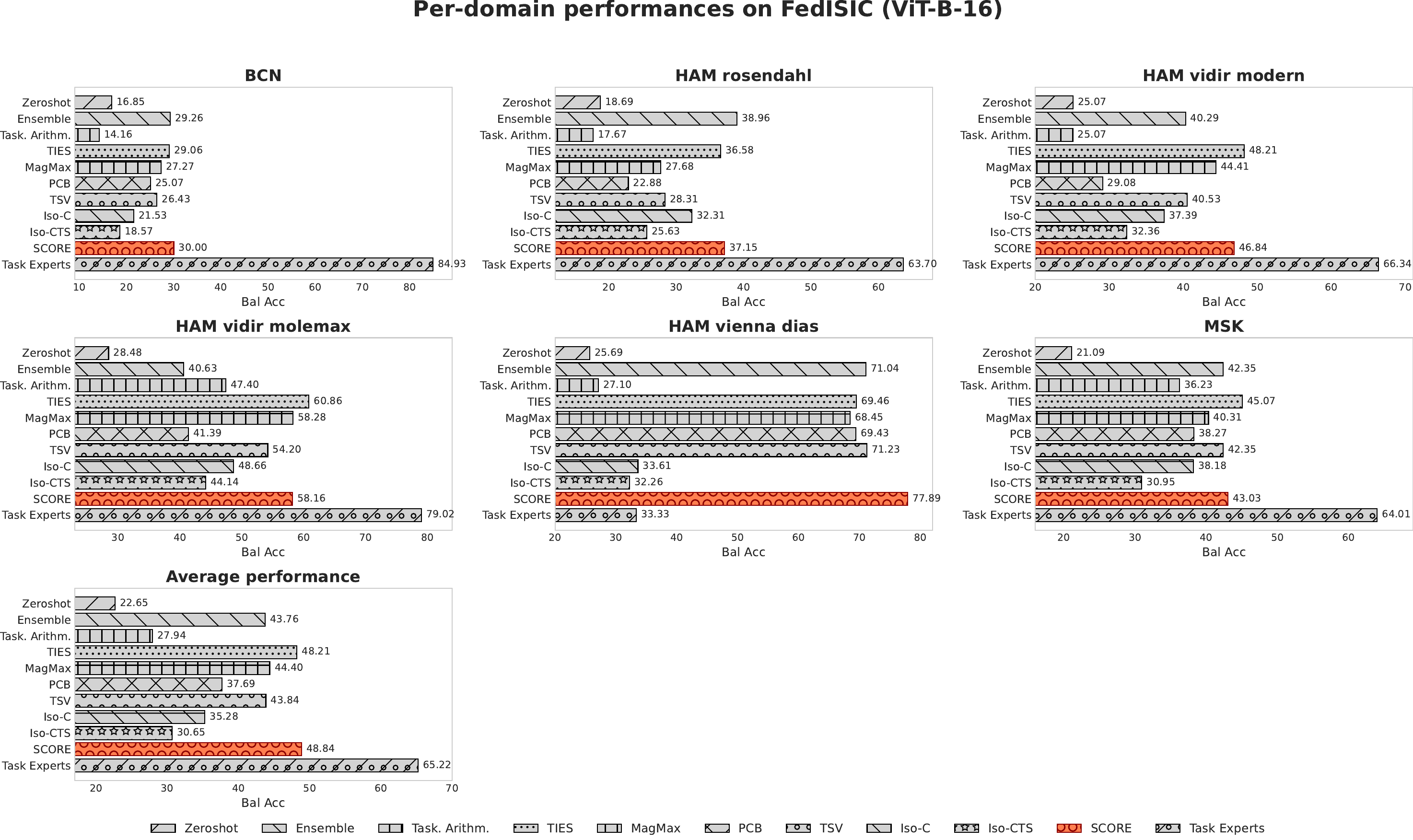}
    \caption{Per-domain results for ViT-B-16 on the FedISIC dataset for each model merging method in our study.}
    \label{fig:vitb16-fedisic}
\end{figure*}

\begin{figure*}[htb!]
    \centering
    \includegraphics[width=\linewidth]{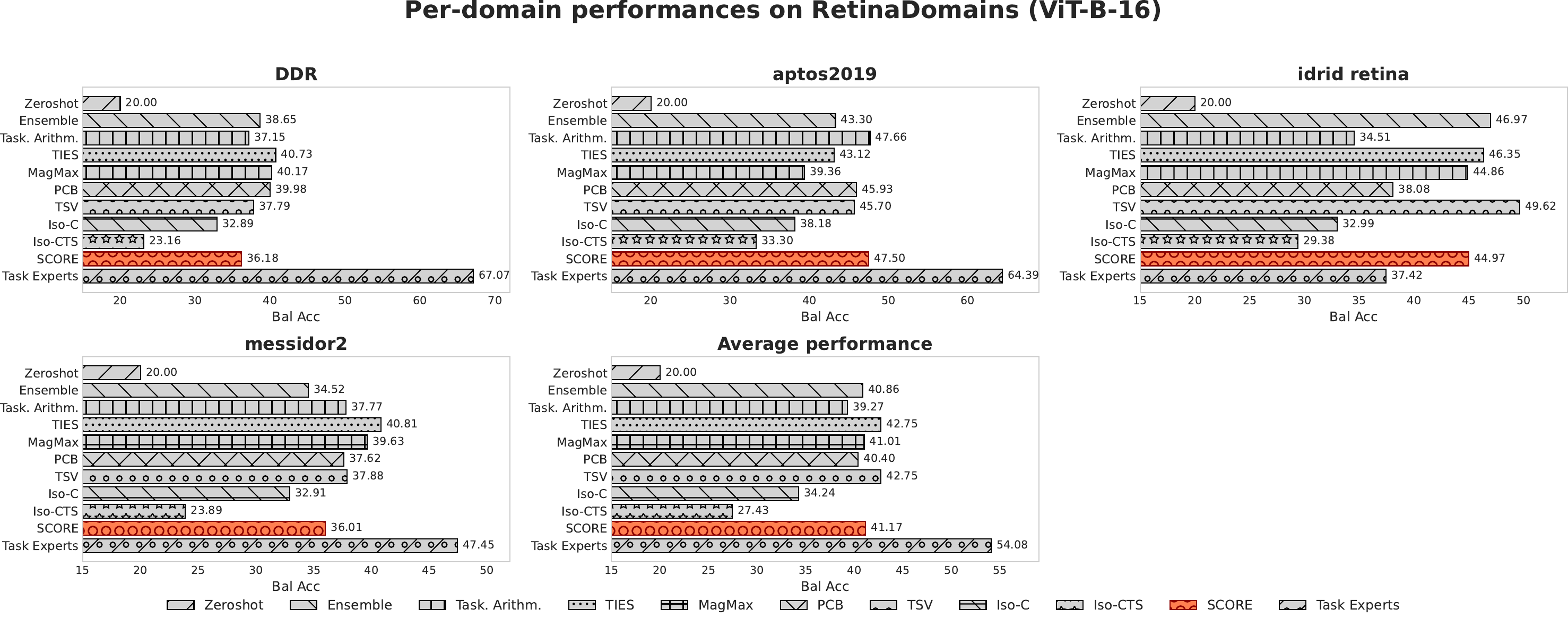}
    \caption{Per-domain results for ViT-B-16 on the RetinaDomains dataset for each model merging method in our study.}
    \label{fig:vitb16-retina}
\end{figure*}

\begin{figure*}[htb!]
    \centering
    \includegraphics[width=\linewidth]{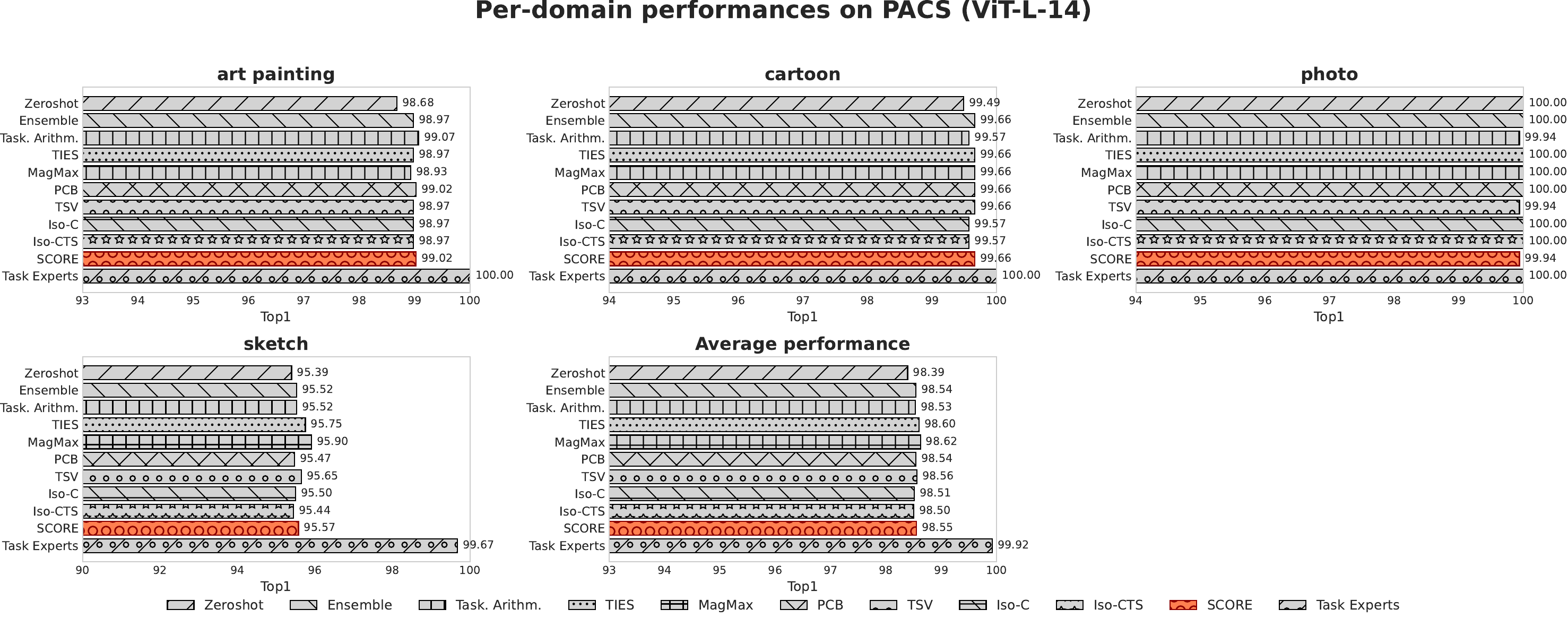}
    \caption{Per-domain results for ViT-L-14 on the PACS dataset for each model merging method in our study.}
    \label{fig:vitl14-pacs}
\end{figure*}

\begin{figure*}[htb!]
    \centering
    \includegraphics[width=\linewidth]{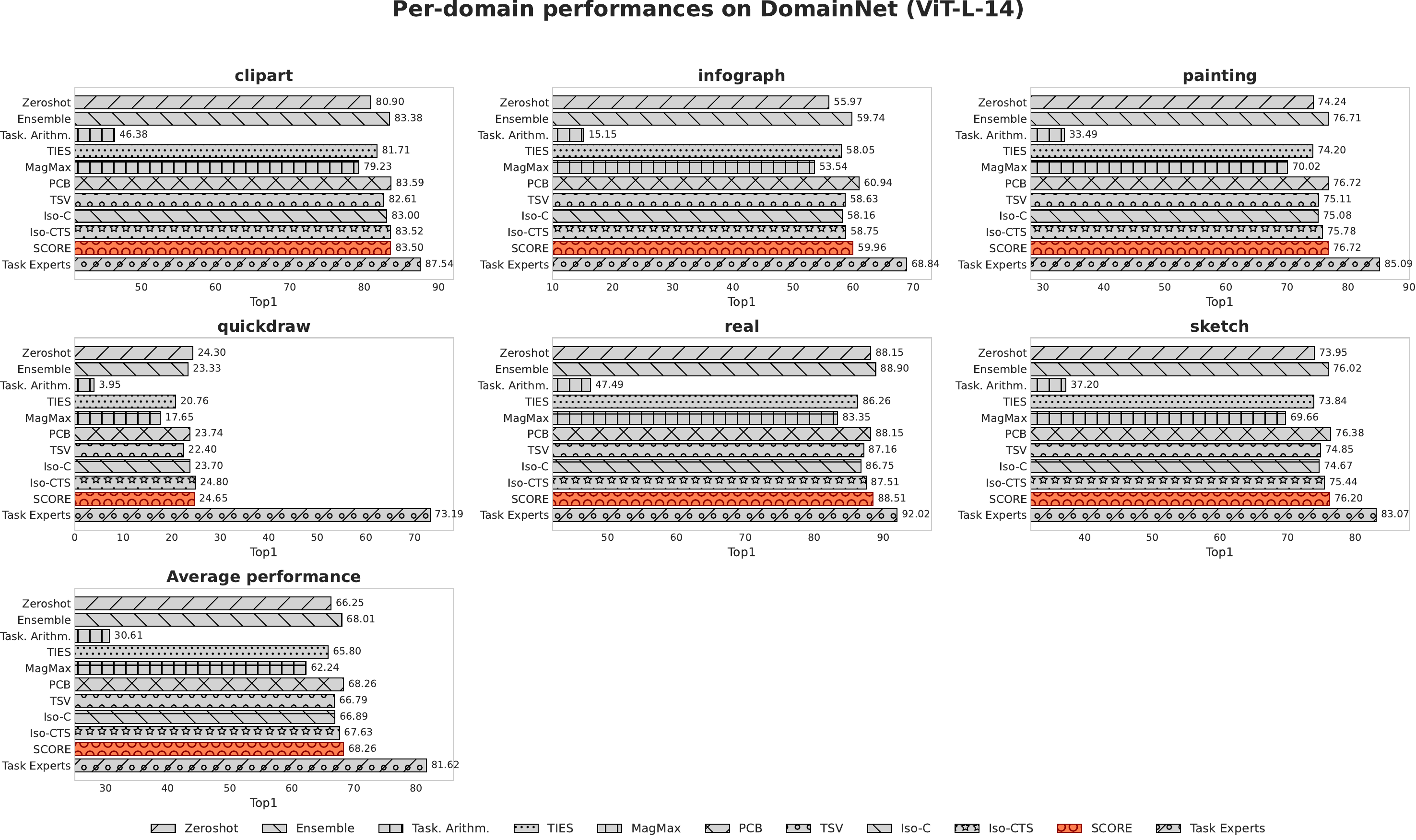}
    \caption{Per-domain results for ViT-L-14 on the DomainNet dataset for each model merging method in our study.}
    \label{fig:vitl14-domainet}
\end{figure*}

\begin{figure*}[htb!]
    \centering
    \includegraphics[width=\linewidth]{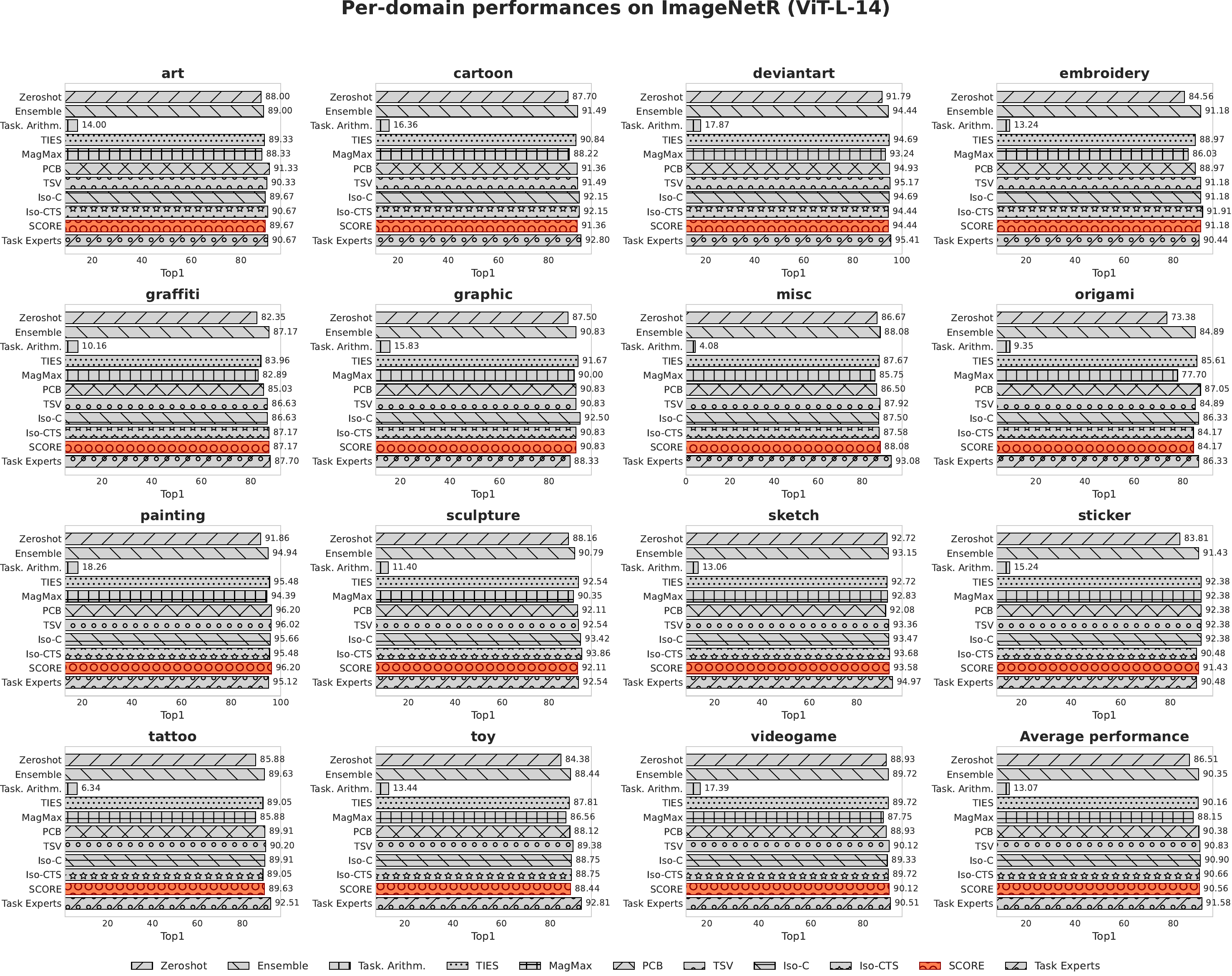}
    \caption{Per-domain results for ViT-L-14 on the ImageNetR dataset for each model merging method in our study.}
    \label{fig:vitl14-imagenetr}
\end{figure*}

\begin{figure*}[htb!]
    \centering
    \includegraphics[width=\linewidth]{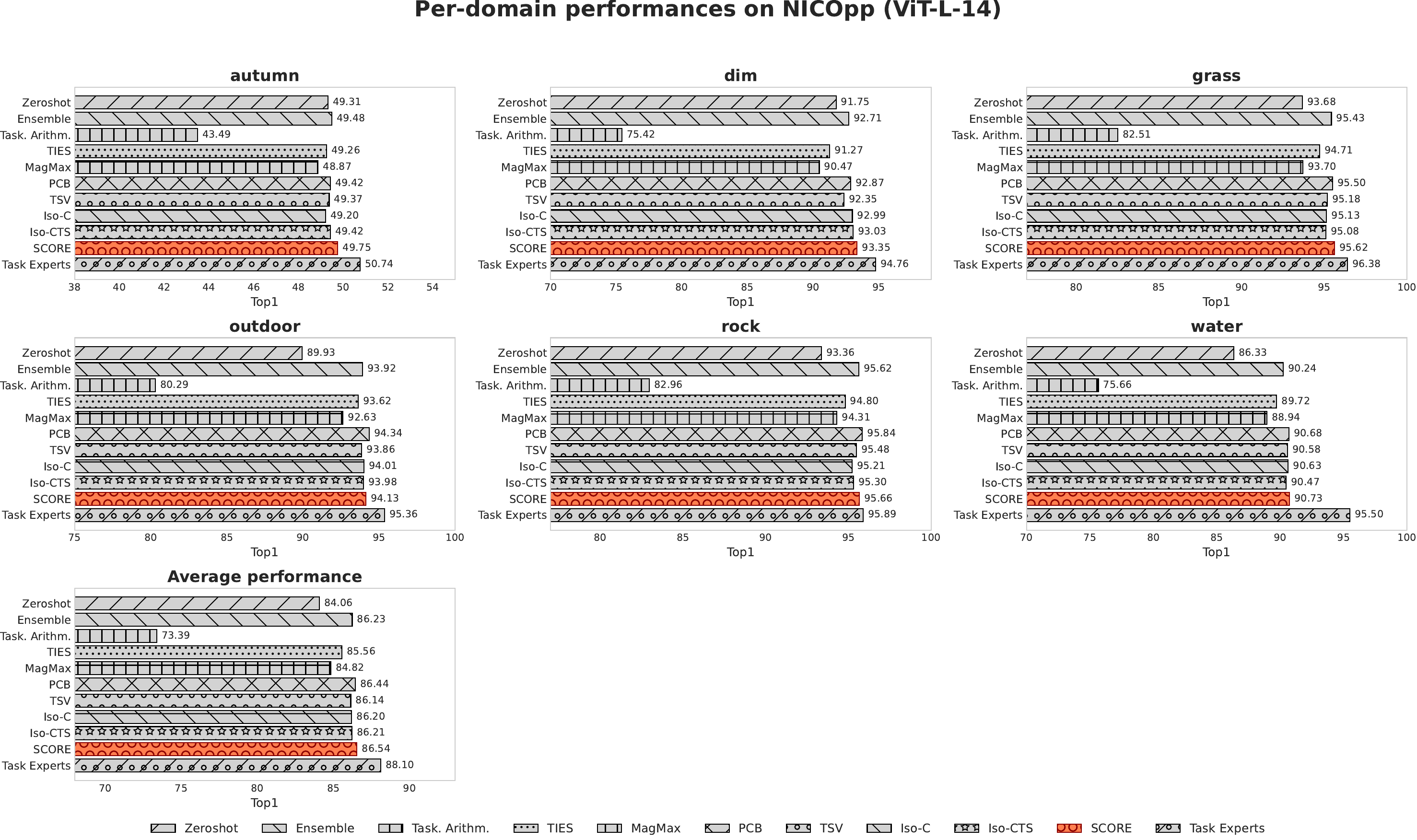}
    \caption{Per-domain results for ViT-L-14 on the NICO++ dataset for each model merging method in our study.}
    \label{fig:vitl14-nicopp}
\end{figure*}

\begin{figure*}[htb!]
    \centering
    \includegraphics[width=\linewidth]{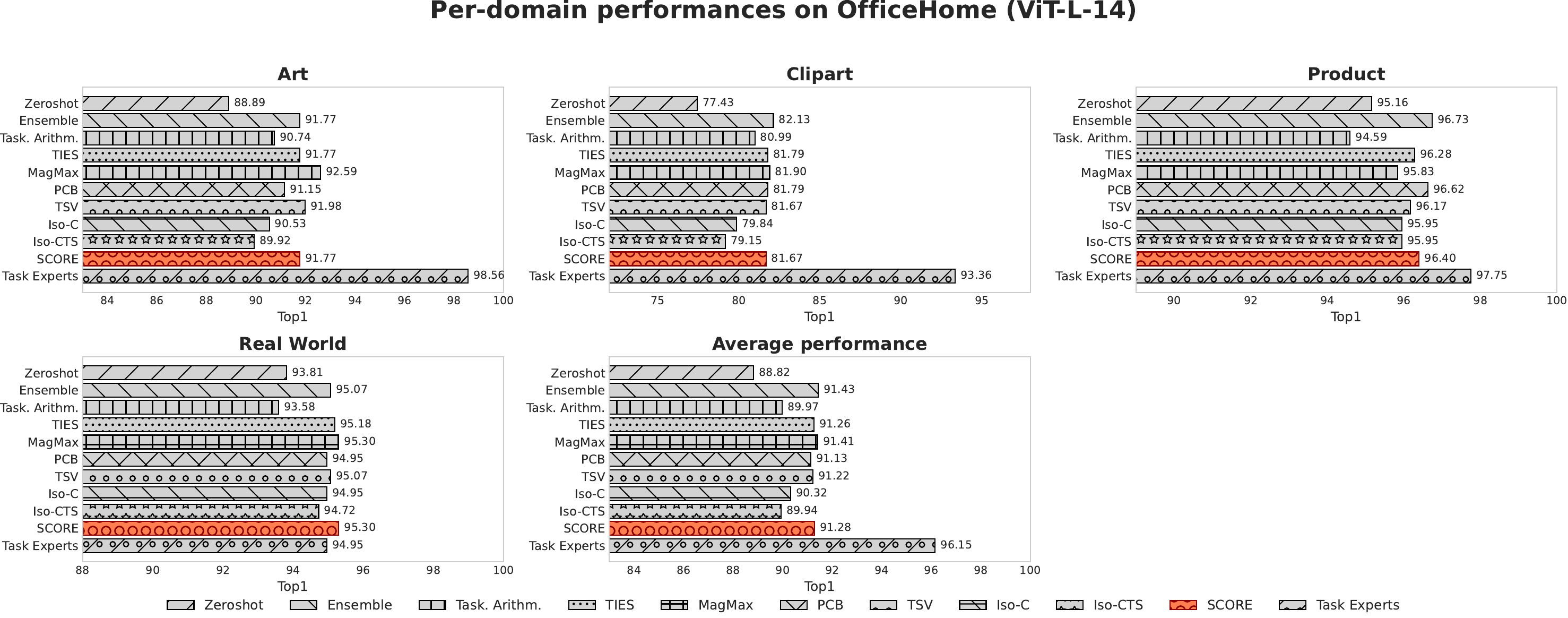}
    \caption{Per-domain results for ViT-L-14 on the OfficeHome dataset for each model merging method in our study.}
    \label{fig:vitl14-officehome}
\end{figure*}

\begin{figure*}[htb!]
    \centering
    \includegraphics[width=\linewidth]{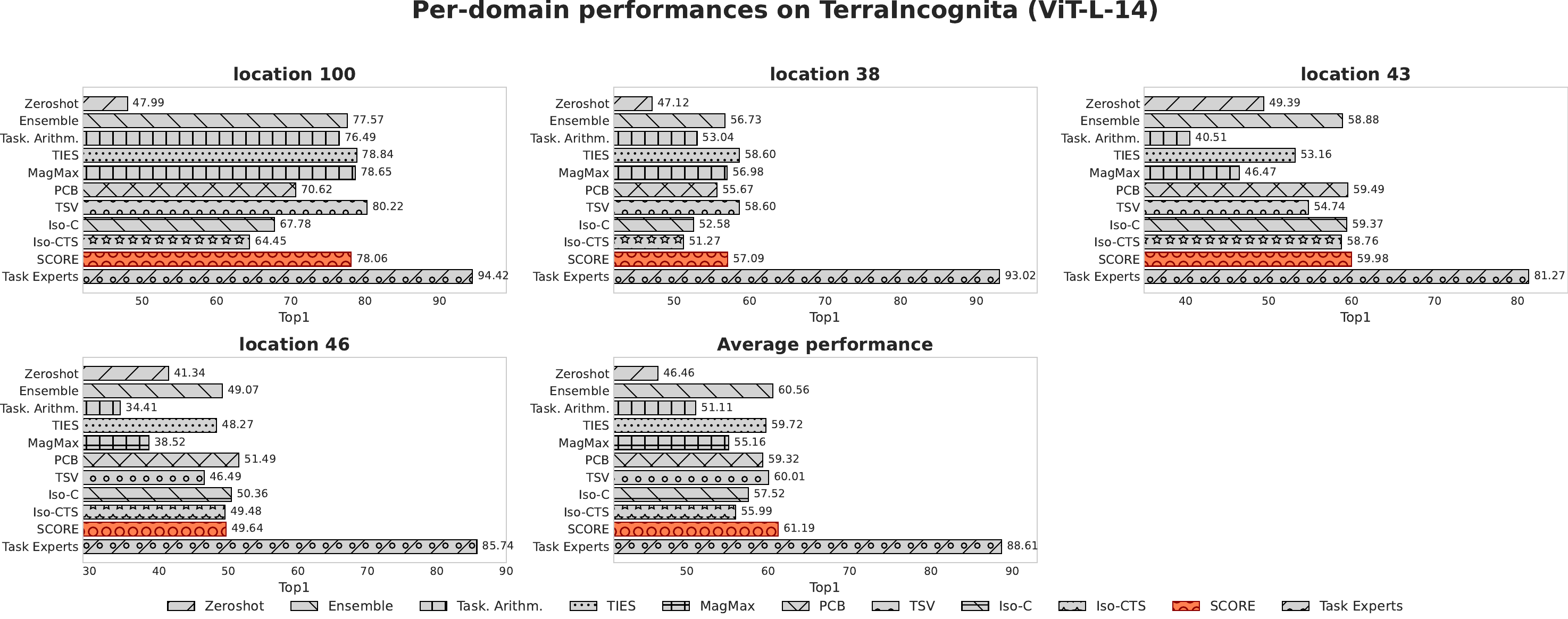}
    \caption{Per-domain results for ViT-L-14 on the TerraIncognita dataset for each model merging method in our study.}
    \label{fig:vitl14-terra}
\end{figure*}

\begin{figure*}[htb!]
    \centering
    \includegraphics[width=\linewidth]{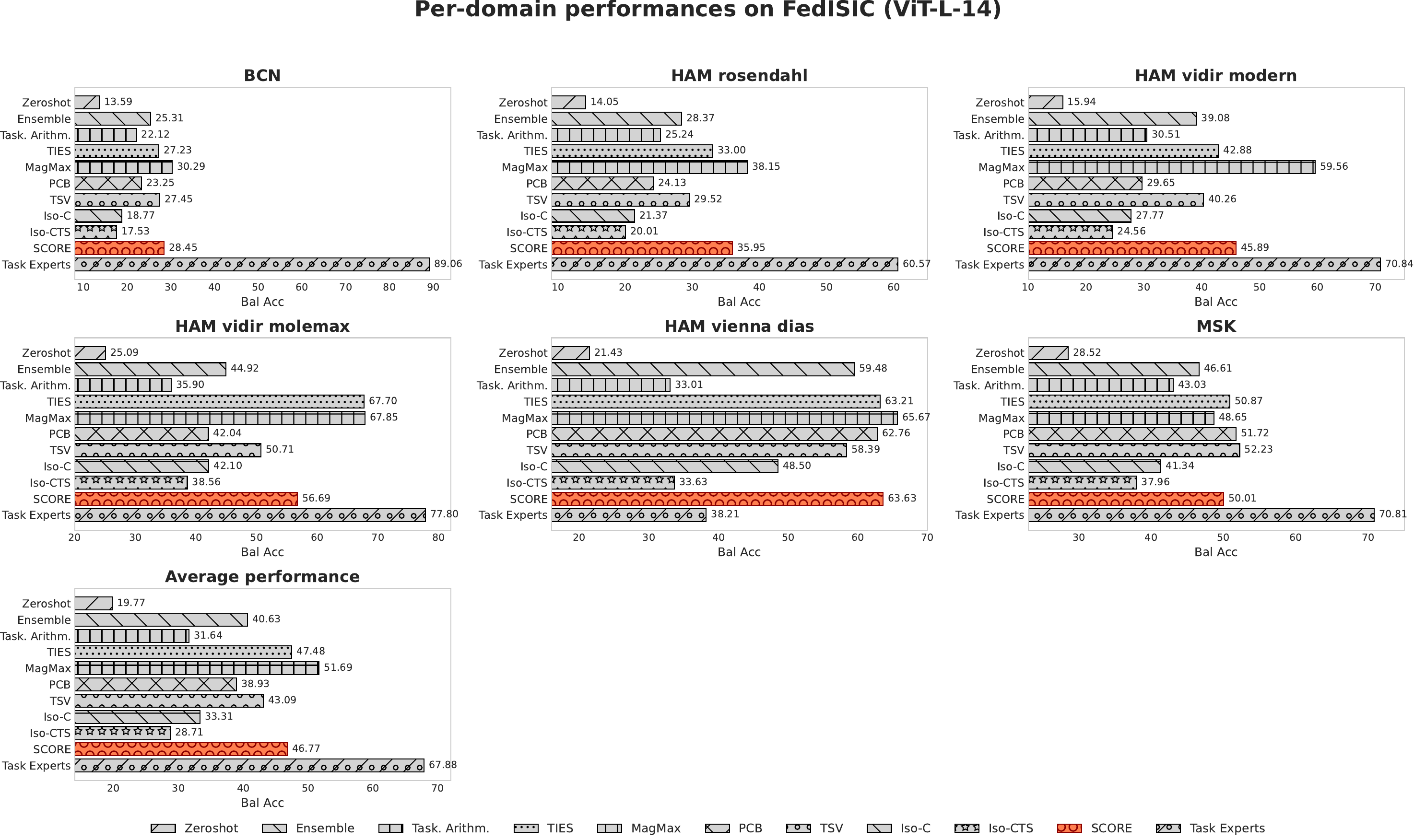}
    \caption{Per-domain results for ViT-L-14 on the FedISIC dataset for each model merging method in our study.}
    \label{fig:vitl14-fedisic}
\end{figure*}

\begin{figure*}[htb!]
    \centering
    \includegraphics[width=\linewidth]{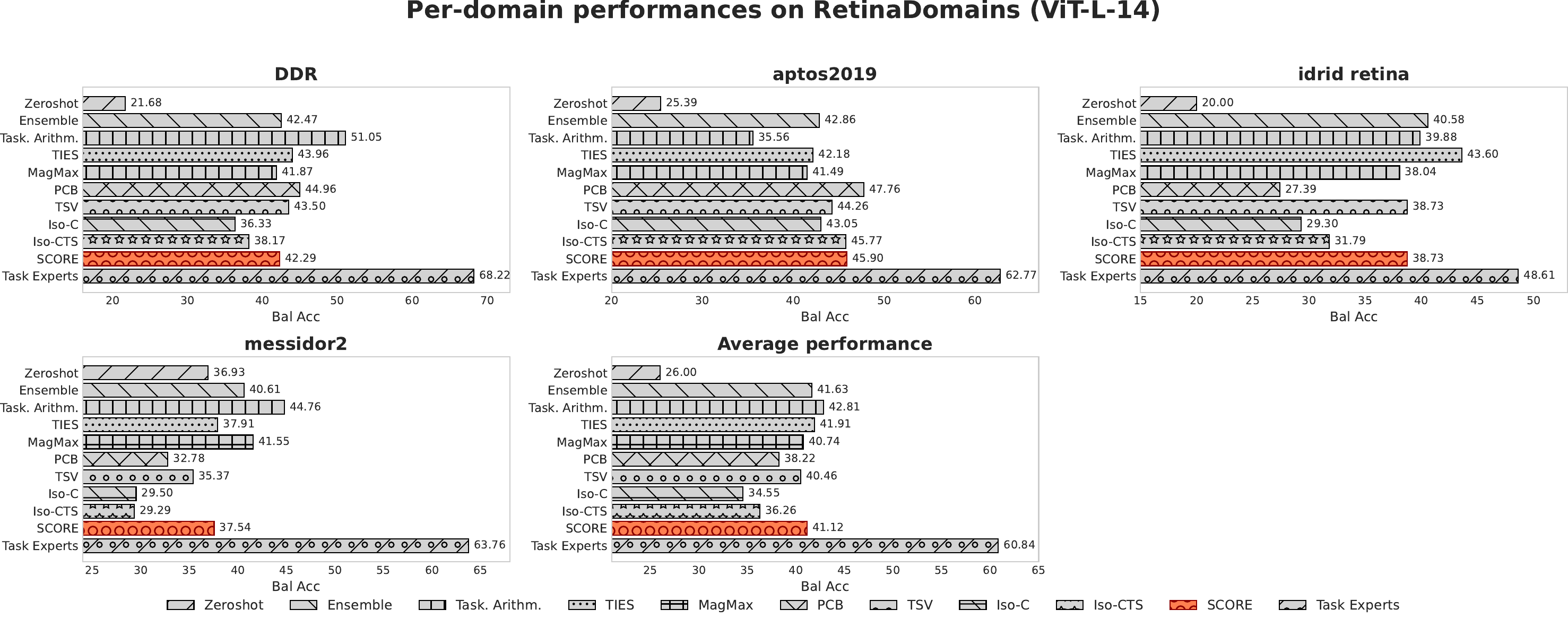}
    \caption{Per-domain results for ViT-L-14 on the RetinaDomains dataset for each model merging method in our study.}
    \label{fig:vitl14-retina}
\end{figure*}

We report per-domain performance of the merged model using the leave-one-domain-out evaluation protocol for ViT-B-32 (Figures \ref{fig:vitb32-pacs}–\ref{fig:vitb32-retina}), ViT-B-16 (Figures \ref{fig:vitb16-pacs}–\ref{fig:vitb16-retina}), and ViT-L-14 (Figures \ref{fig:vitl14-pacs}–\ref{fig:vitl14-retina}). Section \ref{sec:experiments} describes our experimental setup in detail. For completeness, we present results for all model-merging methods we consider in this study, including zero-shot performance, the logits ensemble baseline (see Section \ref{subsec:analysis-ablations}), and the expert performance for each sub-domain.

%  ======= ViT-B-32 results ======

% ===== ViT-B-16 results =======

% ===== ViT-L-14 results =======

\end{document}